\documentclass[lettersize,journal]{IEEEtran}
\usepackage{booktabs}
\usepackage{amsmath,amssymb,amsfonts}

\usepackage{graphicx}
\usepackage[caption=false,font=footnotesize]{subfig}

\usepackage{array}
\usepackage{booktabs}
\usepackage{multirow}
\usepackage{tabularx}
\usepackage{threeparttable}

\usepackage[ruled,vlined,linesnumbered]{algorithm2e}

\usepackage{stfloats}

\usepackage{textcomp}
\usepackage{url}
\usepackage{verbatim}
\usepackage{cite}
\usepackage{xcolor}

\usepackage{hyperref}
\usepackage{cleveref}

\Crefname{figure}{Fig.}{Figs.}
\Crefname{equation}{Eq.}{Eqs.}

\newcommand{\cmark}{$\checkmark$}
\newcommand{\xmark}{$\times$}
\newcommand{\dashmark}{--}

\providecommand{\cmark}{\ding{51}}
\providecommand{\xmark}{\ding{55}}
\providecommand{\dashmark}{--}

\usepackage{flushend}

\usepackage{xcolor}
\newif\ifshowrevisions
\showrevisionsfalse
\newcommand{\rev}[1]{\ifshowrevisions\textcolor{blue}{#1}\else#1\fi}

\begin{document}

\title{Not All Uncertainty Matters: Simulation-in-the-Loop Fast–Slow Reasoning for Decision-Critical Autonomous Driving System}

\author{Jiayi Chen, Shuai Wang$^\dagger$, Guangxu Zhu$^\dagger$,
Derrick Wing Kwan Ng,~\IEEEmembership{Fellow,~IEEE},\\
Chengzhong Xu,~\IEEEmembership{Fellow,~IEEE}, and Kaibin Huang,~\IEEEmembership{Fellow,~IEEE}
\thanks{Part of this work has been presented at the 2025 IEEE/RSJ International Conference on Intelligent Robots and Systems (IROS), Hangzhou, China, in the paper entitled ``Opportunistic Collaborative Planning with Large Vision Model Guided Control and Joint Query-Service Optimization.'' 
Jiayi Chen and Guangxu Zhu are with the Shenzhen Research Institute of Big Data, Shenzhen, China. 
Shuai Wang is with the Center for Cloud Computing, Shenzhen Institutes of Advanced Technology (SIAT), Chinese Academy of Sciences, Shenzhen, China. 
Derrick Wing Kwan Ng is with the School of Electrical Engineering and Telecommunications, The University of New South Wales, Sydney, NSW 2052, Australia. 
Chengzhong Xu is with the State Key Laboratory of Internet of Things for Smart City (SKL-IOTSC), University of Macau, Macau, China. 
Kaibin Huang is with the Department of Electrical and Electronic Engineering, The University of Hong Kong, Hong Kong. 
$^\dagger$ denotes the corresponding authors.}
}

\markboth{IEEE TRANSACTIONS ON MOBILE COMPUTING, ~Vol.~xx, No.~x, July~2026}%
{Shell \MakeLowercase{\textit{et al.}}: A Sample Article Using IEEEtran.cls for IEEE Journals}

\maketitle

\begin{abstract}
Large vision-language models (VLMs) offer powerful open-world perception and reasoning capabilities for autonomous driving. However, their high computational overhead and non-negligible inference latency make continuous cloud-side invocation impractical for real-time operations. This limitation motivates a fast--slow collaborative architecture, where efficient onboard modules perform real-time perception and control tasks while cloud-based models provide selectively high-level reasoning when necessary. The central challenge in such systems is accurately determining when slow cloud reasoning should be involved and allowed to influence time-critical driving decisions. Existing approaches typically rely on perception uncertainty, heuristic triggers, or resource-driven policies, which fail to explicitly evaluate whether resolving a particular uncertainty will actually improve the downstream planning outcome. To address this issue, we propose \textbf{SIGMA}, a simulation-in-the-loop planning-gain modeling and assessment framework for effective task-oriented fast--slow collaboration. The key insight is that not all perception uncertainty equally affect decisions: perception ambiguity when it substantially alters the feasible trajectory set or significantly affects the associated planning cost. Rather than measuring uncertainty solely at the perception level, SIGMA embeds the planner directly into the uncertainty assessment loop to systematically evaluate how plausible scene realizations affect motion-planning outcomes. Specifically, SIGMA constructs candidate scene realizations under semantic and geometric uncertainty, evaluates them through a simulation-in-the-loop planning framework, and estimates the expected reduction in planning cost if the uncertainty is resolved. Building on this formulation, we introduce expected planning gain (EPG), a decision-level metric that quantitatively captures the expected trajectory-level improvement obtainable from cloud reasoning. EPG serves as a unified criterion for managing cloud invocation, cloud-guidance integration, and prioritizing cloud requests under deadline and resource constraints. Extensive experiments in the CARLA simulator demonstrate that SIGMA significantly reduces unnecessary cloud interactions while significantly improving planning performance, driving efficiency, and navigation success in both static and dynamic obstacle scenarios. Compared with fixed-period collaboration, SIGMA reduces unnecessary cloud interactions by 50\%, improves navigation success rate by more than 6\%, and achieves up to 26.2\% reduction in finish time in dynamic scenarios. The implementation of our prediction-time 3D bounding-box-level uncertainty estimator is publicly available at \url{https://github.com/cjychenjiayi/3D_Bbox_uncertainty}.
\end{abstract}

\begin{IEEEkeywords}
Large vision models, uncertainty-aware planning, edge–cloud collaboration, autonomous driving.
\end{IEEEkeywords}

\begin{figure*}[!t]
    \centering
    \scalebox{1.05}[1.00]{%
        \includegraphics[width=0.92\textwidth]{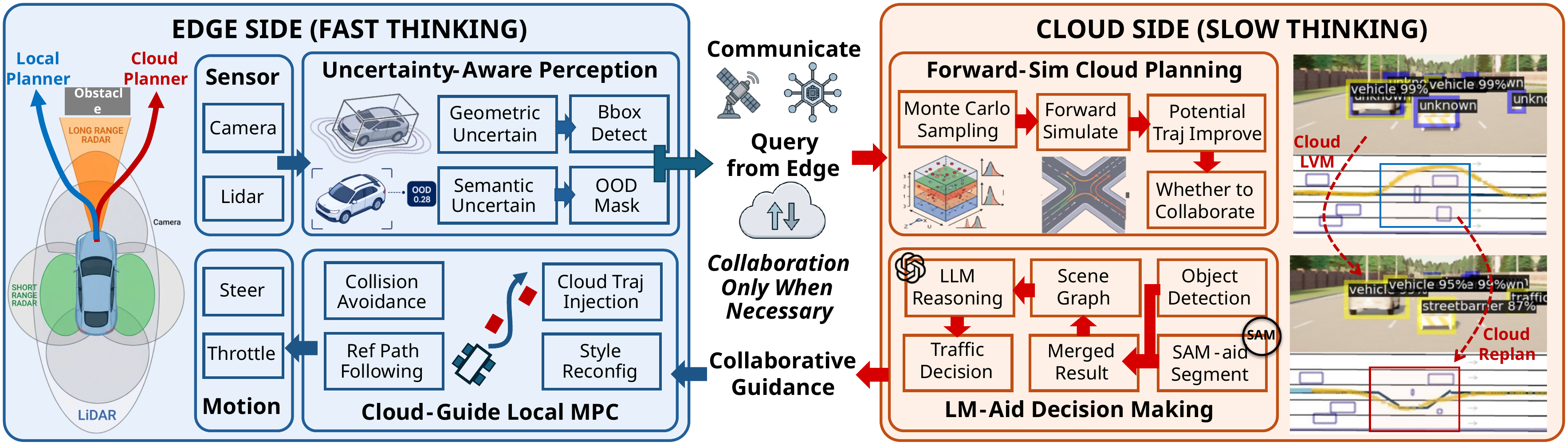}
    }
    \caption{Overview of the proposed planning-gain-guided fast--slow collaborative driving framework.}
    \label{figure:main}
    \vspace{-3mm}
\end{figure*}

\section{Introduction}

Recent advances in large vision, language, and vision--language models have enhanced open-world perception, semantic understanding, and reasoning for autonomous driving~\cite{pan2024vlp,tian2025drivevlm,hwang2024emma}. However, their high computational cost and inference latency make continuous invocation impractical for real-time driving systems, especially in resource-constrained edge--cloud settings. This limitation motivates a \emph{fast--slow} architecture, where efficient onboard modules provide real-time perception, planning and control tasks, while cloud-based models offer slower but more powerful reasoning when necessary~\cite{ichnowski2023fogros2,li2025edge,chen2025joint}. The central challenge is not merely how to leverage large models, but \emph{when slow reasoning should be invoked and allowed to influence real-time driving decisions} under incomplete knowledge of the surrounding world.

Existing autonomous driving systems often determine cloud usage or decision refinement based on perception uncertainty, fixed offloading intervals, communication constraints, or heuristic triggers. Meanwhile, uncertainty-aware perception and planning methods have studied object detection uncertainty, unknown objects, conformal prediction, and robust decision-making under perception ambiguity~\cite{li2024safe,zhang2024multi,timans2024conformal}. Although these methods improve safety and robustness to a certain extent, they do not directly answer a more decision-centric question: \emph{whether resolving the current uncertainty would change or improve the final driving decision}. This limitation reflects a fundamental mismatch between perception-space uncertainty and planning-space decision making. For instance, a perception module may exhibit high uncertainty about distant or irrelevant objects without affecting the planned trajectory, while seemingly confident predictions near critical regions may still result in unsafe or suboptimal driving decisions. Thus, \emph{not all uncertainty matters}; what truly matters is whether alternative plausible scene realizations would induce different downstream planning outcomes.

We argue that the root limitation is not simply inaccurate uncertainty estimation, but \rev{the absence of} an explicit decision\rev{-}evaluation mechanism that systematically considers \emph{multiple plausible realizations of the world}. In real-world driving, the environment is only partially observed, and a single perception output may admit multiple valid semantic and geometric interpretations. Therefore, uncertainty should not be treated only as an intrinsic property of perception, but as a \emph{decision-induced quantity} defined by its effect on planning outcomes. This view is related to forward-simulation-based planning and the emerging ``simulate before execute'' paradigm~\cite{lauri2016planning,ding2022epsilon}. \rev{It is also aligned with the agentic-harness perspective for physical AI systems, where model outputs are mediated by runtime verification, safety, and execution-control modules before being deployed in the physical world.} However, our goal is not to learn a comprehensive generative world model or perform simulation for its own sake. Instead, we leverage forward simulation as a lightweight decision evaluation tool: before allowing slow cloud reasoning to affect the onboard controller, the system estimates whether plausible scene variations would result in meaningful changes in planning outcomes. \rev{In this sense, the proposed simulation-in-the-loop EPG module can be viewed as a planning-grounded verification layer for fast--slow autonomous driving.}

To this end, we introduce \textbf{SIGMA}, a \emph{Simulation-In-the-loop planning-Gain Modeling and Assessment} framework for decision-aware fast--slow collaboration. Rather than relying on a single estimated world, we construct multiple plausible scene realizations by decomposing ambiguity into semantic ``\emph{what}'' and geometric ``\emph{where}'' components. These realizations are then propagated into a simulation-in-the-loop forward simulation, where their downstream effects are evaluated according to the resulting planning costs. By comparing decision outcomes across scene hypotheses, the system can determine whether the ambiguity is decision-relevant and quantifies the potential planning benefit of resolving it. In this formulation, uncertainty is evaluated relying sidely on its \emph{decision impact}, rather than through perception-level proxy metrics.

This perspective naturally leads to a planning-gain-guided view of fast--slow collaboration. The onboard system retains real-time control authority and continuously generates safe and dynamically feasible trajectories, while the cloud provides slower but more reliable semantic reasoning and high-level guidance only when its expected benefit justifies the additional latency and cost. Instead of heuristically switching between onboard and cloud decisions, we introduce a decision-level metric, termed \emph{expected planning gain} (EPG), which estimates the expected improvement in planning quality if additional information or reasoning is incorporated. EPG can be interpreted as a task-specific approximation of the value of information, and it governs both \emph{when to invoke cloud intelligence} and \emph{whether its output should be integrated into local planning}. Under limited computation and communication resources, cloud collaboration is further handled through a planning-gain-aware request prioritization rule that favors requests according to their expected impact on decision outcomes~\cite{chen2025joint,chen2026bridging}.

Overall, the framework integrates uncertainty modeling, simulation-in-the-loop forward simulation, and \emph{planning-gain-guided invocation and integration of cloud guidance into onboard planning}. As illustrated in Fig.~\ref{figure:main}, it shifts the objective of fast--slow collaboration from optimizing proxy metrics such as uncertainty, latency, or invocation frequency to optimizing \emph{decision outcomes under scene variability}.

The main contributions are summarized as follows:
\begin{itemize}
\item \textbf{Decision-coupled uncertainty modeling.}
We formulate perception uncertainty as planning-relevant ambiguity rather than standalone confidence, and decompose it into semantic uncertainty for unknown-object screening and geometric uncertainty for obstacle-level risk modeling. \rev{This yields an explicit semantic--geometric interface from open-world perception to MPC constraints, including a data-driven optimize for semantic screening and probabilistic obstacle inflation for geometric uncertainty.}

\item \textbf{Simulation-in-the-loop planning value estimation.}
We introduce EPG as a decision-level metric that quantifies the value of resolving current ambiguity. \rev{EPG is derived as the expected normalized MPC-cost reduction over plausible scene realizations and estimated via budget-aware Monte Carlo rollouts.} By propagating multiple plausible scene realizations through MPC, EPG identifies uncertainty that truly changes planning outcomes, rather than merely appearing uncertain in perception space.

\item \textbf{Planning-gain-guided fast--slow collaboration.} We design a planning-gain-aware mechanism that adopts EPG to regulate cloud invocation, cloud-guided MPC reference switching, and deadline-aware prioritization. The cloud acts as selective guidance, while real-time feasibility and safety remain by onboard MPC.

\item \textbf{End-to-end validation in open-world driving scenarios.}
We implement SIGMA in CARLA and evaluate it under static and dynamic unknown-obstacle scenarios. Compared with fixed-period collaboration, SIGMA reduces unnecessary cloud interactions by 50\%, improves navigation success rate by more than 6\%, and achieves up to 26.2\% reduction in finish time in dynamic scenarios.
\end{itemize}

\emph{Symbol Notation}: Throughout the paper, scalars are represented by italic letters, vectors or lists by bold lowercase letters, matrices by bold uppercase letters, and sets or scene representations by calligraphic uppercase letters.

\begin{table*}[!t]
    \centering
    \caption{Feature-level comparison with representative related works.}
    \label{tab:related_work_comparison}
    \scriptsize
    \setlength{\tabcolsep}{3.0pt}
    \renewcommand{\arraystretch}{1.15}

    \begin{tabularx}{\textwidth}{p{0.2\textwidth}*{9}{>{\centering\arraybackslash}X}}
        \toprule
        \multirow{2}{*}{\textbf{Work / Ref.}}
            & \multicolumn{3}{c}{\textbf{Uncertainty Modeling}}
            & \multicolumn{3}{c}{\textbf{Capability / constraints}}
            & \multicolumn{3}{c}{\textbf{Planning-level decision awareness}} \\
        \cmidrule(lr){2-4}
        \cmidrule(lr){5-7}
        \cmidrule(lr){8-10}

            & \begin{tabular}[c]{@{}c@{}}Semantic\\uncertain\end{tabular}
            & \begin{tabular}[c]{@{}c@{}}Geometric\\uncertain\end{tabular}
            & \begin{tabular}[c]{@{}c@{}}Plan\\impact\end{tabular}
            & \begin{tabular}[c]{@{}c@{}}Real\\time\end{tabular}
            & \begin{tabular}[c]{@{}c@{}}Open-world\\capacity\end{tabular}
            & \begin{tabular}[c]{@{}c@{}}Resource\\aware\end{tabular}
            & \begin{tabular}[c]{@{}c@{}}Planner\\integration\end{tabular}
            & \begin{tabular}[c]{@{}c@{}}Collaboration\\criterion\end{tabular}
            & \begin{tabular}[c]{@{}c@{}}Planning-gain\\modeling\end{tabular} \\
        \midrule

        Xiao \emph{et al.} (TMC'24)~\cite{xiao2024occlusionfree}
            & \dashmark & \dashmark & \dashmark
            & High & Low & \cmark
            & \dashmark & Resource & \dashmark \\

        Ye \emph{et al.} (TMC'24)~\cite{ye2024accuracyaware}
            & \cmark & \xmark & \xmark
            & High & Low & \cmark
            & \dashmark & Accuracy/resource & \dashmark \\

        Fang \emph{et al.} (TMC'24)~\cite{fang2024pacp}
            & \dashmark & \dashmark & \dashmark
            & High & Low & \cmark
            & \dashmark & Perception priority & \dashmark \\

        Lin \emph{et al.} (TMC'25)~\cite{lin2025fabrication}
            & \cmark & \xmark & \xmark
            & Medium & Low & \cmark
            & \dashmark & Security & \dashmark \\

        Liang \emph{et al.} (TMC'26)~\cite{liang2026fullperception}
            & \dashmark & \dashmark & \dashmark
            & High & Medium & \cmark
            & \dashmark & Blind spot & \dashmark \\

        Yang \emph{et al.} (TMC'26)~\cite{yang2026avrin}
            & \dashmark & \dashmark & \dashmark
            & High & Low & \cmark
            & \dashmark & Resource & \dashmark \\

        Zhou \emph{et al.} (TMC'26)~\cite{zhou2026ecseg}
            & \dashmark & \dashmark & \dashmark
            & Medium & Low & \cmark
            & \dashmark & Energy/carbon & \dashmark \\

        Tang \emph{et al.} (TMC'22)~\cite{tang2022drl}
            & \xmark & \xmark & \xmark
            & Medium & Low & \cmark
            & \dashmark & Edge-load uncertainty & \dashmark \\

        Xia \emph{et al.} (TMC'24)~\cite{xia2024uncertainties}
            & \xmark & \xmark & \xmark
            & Medium & Low & \cmark
            & \dashmark & Task/network uncertainty & \dashmark \\

        Cheng \emph{et al.} (TMC'25)~\cite{cheng2025uncertain}
            & \xmark & \xmark & \xmark
            & Medium & Low & \cmark
            & \dashmark & Network/QoS uncertainty & \dashmark \\

        Xu \emph{et al.} (TMC'26)~\cite{xu2026multifactorial}
            & \xmark & \xmark & \xmark
            & Medium & Low & \cmark
            & \dashmark & Multifactor uncertainty & \dashmark \\

        Li \emph{et al.} (TIV'24)~\cite{li2024safe}
            & \xmark & \cmark & \cmark
            & Medium & Low & \dashmark
            & MPC & Perception uncertainty & \xmark \\

        Li \emph{et al.} (TMECH'24)~\cite{li2025edge}
            & \dashmark & \dashmark & \dashmark
            & High & Low & \cmark
            & MPC & Resource & \dashmark \\

        Tian \emph{et al.} (ICRA'24)~\cite{tian2025drivevlm}
            & \dashmark & \dashmark & \dashmark
            & Low & High & \dashmark
            & \dashmark & Semantic & \dashmark \\

        \textbf{SIGMA (ours)}
            & \cmark & \cmark & \cmark
            & High & High & \cmark
            & MPC & Value & \cmark \\

        \bottomrule
    \end{tabularx}

    \begin{flushleft}
        \scriptsize
        \textit{Note:} In the uncertainty columns, \cmark means that the item is explicitly modeled,
        \xmark means that the work considers uncertainty but not this item, and \dashmark means that
        uncertainty is not a main modeling target or the item is not applicable. QoS denotes quality of service.
    \end{flushleft}

    \vspace{-4.5mm}
\end{table*}

\section{Related Work}

\subsection{Uncertainty Estimation in Autonomous Driving Perception}

Reliable uncertainty estimation is crucial for autonomous driving, as perception errors can propagate to downstream planning and compromise safety. Existing works primarily focus on uncertainty at the perception level, including robust detection under adverse conditions~\cite{jin2023robust}, probabilistic object detection with localization uncertainty~\cite{choi2019gaussian}, unknown-object discovery~\cite{wong2020identifying,liang2023unknown}, and uncertainty calibration with statistical guarantees~\cite{timans2024conformal,huseljic2025systematic,doula2025conformal}. These approaches improve the detection reliability and provide more accurate characterization of uncertainty in open-world perception.

However, such methods treat uncertainty mainly as an intrinsic property of perception outputs, focusing on robustness, calibration, or unknown-instance awareness. They do not explicitly address whether a given uncertainty is \emph{decision-relevant}, i.e., whether resolving it would change the downstream planning outcome. In contrast, our work adopts a task-oriented perspective, where uncertainty is evaluated based on its impact on planning rather than its magnitude alone.

\subsection{Motion Planning Under Perception Uncertainty}

Another line of work incorporates uncertainty into motion planning and decision making. Classical approaches include formal verification and robust control~\cite{xu2019quantitative,yang2023towards}, as well as planning frameworks that explicitly account for uncertainty through chance-constrained optimization, MPC variants, or integrated perception--planning pipelines~\cite{li2024conformalmpc,li2024safe,mohseni2018deep,ding2022epsilon,han2023rda,zhang2024multi}. Related studies have investigated distributed or collaborative planning under uncertainty~\cite{mohseni2021distributed,zhang2022collision}. More broadly, some prior work on informative planning and forward-simulation-based exploration explicitly accounts for uncertainty when evaluating and selecting candidate actions~\cite{lauri2016planning,popovic2024aipp,kim2025cipp}.

These works primarily aim to improve safety and robustness under uncertainty, typically treating uncertainty as a factor to be handled conservatively once it appears. What remains underexplored is whether all uncertainty should be treated equally, and more importantly, whether resolving a particular uncertainty would \emph{improve the optimal decision}. Our work addresses this gap by explicitly quantifying the planning gain of uncertainty resolution through simulation-in-the-loop evaluation, enabling the system to distinguish between decision-critical and decision-irrelevant uncertainty.

\subsection{Fast--Slow Collaborative Driving}

Recent autonomous driving systems increasingly adopt a fast--slow collaborative paradigm. On one hand, cloud- and edge-assisted frameworks enable vehicles to offload computation tasks and collaborate with external resources~\cite{liu2023ros,tanwani2020rilaas,ichnowski2023fogros2,li2025edge,gao2024vrcp,chen2025joint}. On the other hand, large models have been introduced to substantially enhance semantic understanding and high-level decision making in autonomous driving, including language-guided planning and driving agents~\cite{pan2024vlp,xu2024drivegpt4,cui2025drivemlm,wang2023bevgpt,hu2024agentscodriver,sha2023languagempc,wen2024dilu,tian2025drivevlm,hwang2024emma}. These studies demonstrate the potential of coupling lightweight onboard modules for real-time perception, planning, and control with more capable cloud-side reasoning in complex scenarios. However, existing collaborative systems typically rely on heuristic triggers, predefined pipelines, or capability-driven designs to decide when cloud or large-model reasoning should be invoked. Likewise, large-model-based driving methods mainly focus on improving reasoning capability, but often assume that the generated guidance, once available, should directly affect downstream decision making. As a result, slow reasoning may be unnecessarily triggered even when it brings little planning benefit, and its outputs may be incorporated without explicitly assessing their impact on the final driving decision.

In contrast, our framework introduces a decision-centric perspective for fast--slow collaboration: both the invocation of cloud reasoning and its integration into local planning are governed by a unified planning-gain criterion. By evaluating whether uncertainty resolution leads to measurable improvements to planning performance, the system allocates both computational resources and decision authority.

\subsection{Positioning of Our Work}

Existing studies on perception uncertainty, uncertainty-aware planning, and collaborative driving are largely developed along separate directions: perception works estimate uncertainty, planning works handle uncertainty, and collaborative or large-model systems enhance reasoning capability. What remains underexplored is a unified decision-oriented perspective, where both uncertainty resolution and external reasoning are governed by their expected impact on downstream planning. Our work adopts this view by evaluating uncertainty and fast--slow collaboration through EPG, a planning-level criterion.

Table~\ref{tab:related_work_comparison} further positions SIGMA against representative studies in perception uncertainty, uncertainty-aware planning, collaborative perception, and edge--cloud service scheduling. Existing collaborative and resource-aware systems provide strong mechanisms for computation offloading, communication scheduling, perception enhancement, or service-cost optimization, including under uncertain network or workload conditions. However, these systems generally optimize proxy objectives such as perception quality, resource usage, latency, or service cost, rather than the downstream planning gain obtained by resolving uncertainty. SIGMA differs by using simulation-in-the-loop planning cost to decide both when uncertainty deserves slow reasoning and when the resulting guidance should affect real-time control.

\section{System Model and Problem Formulation}
\label{sec:system_overview}
As illustrated in Fig.~\ref{figure:main}, we consider a fast--slow collaborative autonomous driving system consisting of an onboard edge side and a cloud side. The edge side acts as the fast-thinking component. It receives multi-modal observations from onboard sensors, such as cameras and LiDAR, and performs real-time perception, planning, and control. Its uncertainty-aware perception module constructs a local planning-oriented scene representation from the current observation, including detected objects, geometric uncertainty of obstacle states, and semantic uncertainty for unknown-object awareness. Based on this local representation, the onboard MPC planner performs reference-path following and collision avoidance, and outputs low-level control commands such as steering and throttle input.

The cloud side acts as the slow-thinking module. When collaboration is requested, it performs forward simulation over multiple plausible scene realizations to estimate whether resolving the current uncertainty would improve the downstream trajectory. It can also refine perception and generate high-level driving guidance using large vision or vision--language models, aided by object detection, segmentation, scene-graph construction, and traffic reasoning. The cloud output is not directly used as a low-level control command; instead, it is returned as refined scene information or high-level trajectory guidance and incorporated into the onboard MPC planner. Therefore, real-time control authority and safety enforcement remain on the edge side, while the cloud is selectively used only when slow reasoning is expected to improve the planning outcome before the decision deadline.

We consider the trajectory decision problem of an autonomous vehicle assisted by a large vision or vision--language model. At each time step $t$, the ego vehicle observes the surrounding environment and plans a future trajectory $\tau_t$. The goal is to select a safe, efficient, and dynamically smooth trajectory under uncertain perception.
In particular, let $\mathcal{C}_t$ denote a planning-oriented scene configuration, while $c_{q,t}$ denotes a scalar detection confidence score.
Under ideal perception and reasoning, trajectory planning can be formulated as the minimization of a trajectory-level cost function:
\begin{align}
\boldsymbol{\tau}_t^\star=
\operatorname*{arg\,min}_{\boldsymbol{\tau}_t \in \mathcal{T}_t}
 \ 
J(\boldsymbol{\tau}_t), 
\label{eq:ideal_planning}
\end{align}
where $\boldsymbol{\tau}_t$ is the trajectory list and $\mathcal{T}_t$ is the feasible trajectory set at time step $t$. 
The cost function $J(\tau_t)$ needs to account for trajectory length $L(\tau_t)$, travel or completion time $T(\tau_t)$, smoothness cost $S(\tau_t)$, and safety or collision-risk cost $R(\tau_t)$. 
One widely-adopted weighted-sum formulation is \cite{han2024efficient}:
\begin{align}
J(\boldsymbol{\tau}_t)=w_l L(\boldsymbol{\tau}_t) + w_T T(\boldsymbol{\tau}_t) +w_s S(\boldsymbol{\tau}_t)+w_r R(\boldsymbol{\tau}_t),
\end{align}
where parameters $(w_l, w_T, w_s, w_r)$ are non-negative weights that balance the relative importance of each objective. 


In real-world driving scenarios, however, the onboard system has only partial and uncertain observations of the environment. Unknown or ambiguous objects may be incorrectly classified, conservatively treated as obstacles, or missed by the local perception module. Consequently, the onboard planner may optimize over an inaccurate scene representation leads to a conservative, unsafe, or detouring suboptimal trajectory.

This motivates a fast--slow architecture. The onboard module serves as the fast system, providing real-time perception, planning, and control. The cloud-side model serves as the slow system and provides more accurate semantic recognition and high-level reasoning for uncertain or unknown objects. Nevertheless, invoking the large model introduces communication delay and computation cost. Therefore, continuous invocation is inefficient and may even be detrimental when the response arrives too late to affect the current planning decision.

Let $\mathbf{o}_t$ denote the local observation at time step $t$. The onboard system maps $\mathbf{o}_t$ to a planning-oriented local scene representation $\mathcal{C}_t^{\mathrm{loc}}$, while the large model, once invoked, returns a refined scene representation $\mathcal{C}_t^{\mathrm{lm}}$. 
Accordingly, the perfect feasible trajectory set $\mathcal{T}$ becomes $\mathcal{T}(\mathcal{C}_t^{\mathrm{loc}})$ and $\mathcal{T}(\mathcal{C}_t^{\mathrm{lm}})$ under local and large-model-refined scene representations, respectively.
As such, the local and large-model-assisted planning results are redefined as
\begin{align}
\left\{
\boldsymbol{\tau}_t^{\mathrm{loc}},
\boldsymbol{\tau}_t^{\mathrm{lm}}
\right\}
=
\left
\{
\operatorname*{arg\,min}_{\boldsymbol{\tau}_t \in \mathcal{T}(\mathcal{C}_t^{\mathrm{loc}})}
 \ 
J(\boldsymbol{\tau}_t), 
\operatorname*{arg\,min}_{\boldsymbol{\tau}_t \in \mathcal{T}(\mathcal{C}_t^{\mathrm{lm}})}
 \ 
J(\boldsymbol{\tau}_t)
\right
\}.
\label{eq:local_lm_decision}
\end{align}
In this work, we assume that once the cloud result is returned before the decision deadline, the large model provides more accurate semantic recognition and reliable high-level decision guidance. Each invocation, however, incurs a response delay $\Delta_t^{\mathrm{lm}}$ and an invocation cost $\kappa^{\mathrm{lm}}$.

To describe whether the large model should be invoked, we introduce $\alpha_t\in\{0,1\}$, where $\alpha_t=1$ indicates cloud invocation. 
Let 
$J_t^{\mathrm{lm}}=J(\tau_t^{\mathrm{lm}})$ and $J_t^{\mathrm{loc}}=J(\tau_t^{\mathrm{loc}})$.
Since $J_t^{\mathrm{lm}}$ and the realized delay $\Delta_t^{\mathrm{lm}}$ are unknown before cloud execution, we need an oracle formulation:
\begin{equation}
\begin{aligned}
\min_{\{\alpha_t\}}\quad
&
\sum_{t=0}^{T}
\left[
(1-\alpha_t)J_t^{\mathrm{loc}}
+
\alpha_t
\left(
J_t^{\mathrm{lm}}
+\lambda \kappa^{\mathrm{lm}}
+\rho D(\Delta_t^{\mathrm{lm}})
\right)
\right] \\
\mathrm{s.t.}\quad
&
\alpha_t\in\{0,1\},\quad t=0,\ldots,T,\\
&
\alpha_t
\left(
\Delta_t^{\mathrm{lm}}-\Delta_t^{\mathrm{ddl}}
\right)
\leq 0,\quad t=0,\ldots,T .
\end{aligned}
\label{eq:horizon_optimization}
\end{equation}
Here, $D$ is the delay cost function, and $\Delta_t^{\mathrm{ddl}}$ is the decision deadline. The last constraint means that, if $\alpha_t=1$, the cloud response must satisfy $\Delta_t^{\mathrm{lm}}\leq\Delta_t^{\mathrm{ddl}}$; otherwise no cloud response is required. 

This equation \eqref{eq:horizon_optimization} is an oracle benchmark rather than an online policy, because $J_t^{\mathrm{lm}}$ and $\Delta_t^{\mathrm{lm}}$ are unavailable before cloud execution.
We therefore define the \emph{Expected Planning Gain} (EPG) as the expected normalized reduction in MPC planning cost caused by resolving the current planning-relevant uncertainty. Let $\mathcal{C}_t^{\mathrm{base}}$ be the conservative scene configuration from local perception, and let $\mathcal{C}_t(\boldsymbol{\xi}_t)$ be a plausible uncertainty-resolved configuration with $\boldsymbol{\xi}_t\sim p_t(\boldsymbol{\xi}\mid \mathbf{o}_t)$. Then
\begin{align}
\mathrm{EPG}_t
&=
\mathbb{E}_{\boldsymbol{\xi}_t\sim p_t(\boldsymbol{\xi}\mid \mathbf{o}_t)}
\left[
\underbrace{\frac{
J_t(\mathcal{C}_t^{\mathrm{base}})-J_t(\mathcal{C}_t(\boldsymbol{\xi}_t))
}{
J_t(\mathcal{C}_t^{\mathrm{base}})
}}_{:=\Delta J_t(\boldsymbol{\xi}_t)}
\right].
\label{eq:epg_decision_pipeline}
\end{align}
We also define a benefit factor 
$\tilde{g}_t=
\mathrm{EPG}_t
-\lambda \kappa^{\mathrm{lm}}
-\rho D(\widehat{\Delta}_t^{\mathrm{lm}})$,
where $\widehat{\Delta}_t^{\mathrm{lm}}$ is the \rev{predicted} response delay. The cloud is invoked only when the estimated planning benefit outweighs the invocation and delay costs.
Therefore, a necessary \rev{benefit-side} condition for cloud invocation is 
\rev{$\tilde{g}_t>\theta_g$}, \rev{where $\theta_g$ denotes the benefit-factor threshold.}
\rev{The complete online invocation rule further combines this benefit-side condition with deadline feasibility, as described in Section~\ref{subsec:cloud_guided_mpc}.}

\section{Decision-Oriented Uncertainty}
\label{sec:uncertainty_modeling}

To support planning-gain-guided collaboration between onboard and cloud intelligence, uncertainty should be in a form that is directly relevant to downstream decisions. Rather than treating uncertainty as a generic property of perception outputs, we view it as task-dependent and defined by its impact on planning. Some uncertainties directly affect trajectory optimization, while others mainly indicate possible perception failures with limited influence on the optimal motion.

We therefore propose a decision-oriented uncertainty representation that decomposes scene uncertainty into semantic and geometric components. Semantic uncertainty captures ambiguity in object identity and serves as a screening cue for acquiring additional information, such as invoking cloud assistance. Geometric uncertainty captures ambiguity in object localization and spatial extent, directly influencing collision risk, and trajectory optimization.

\begin{figure}[!t]
    \centering
    \includegraphics[width=0.92\columnwidth]{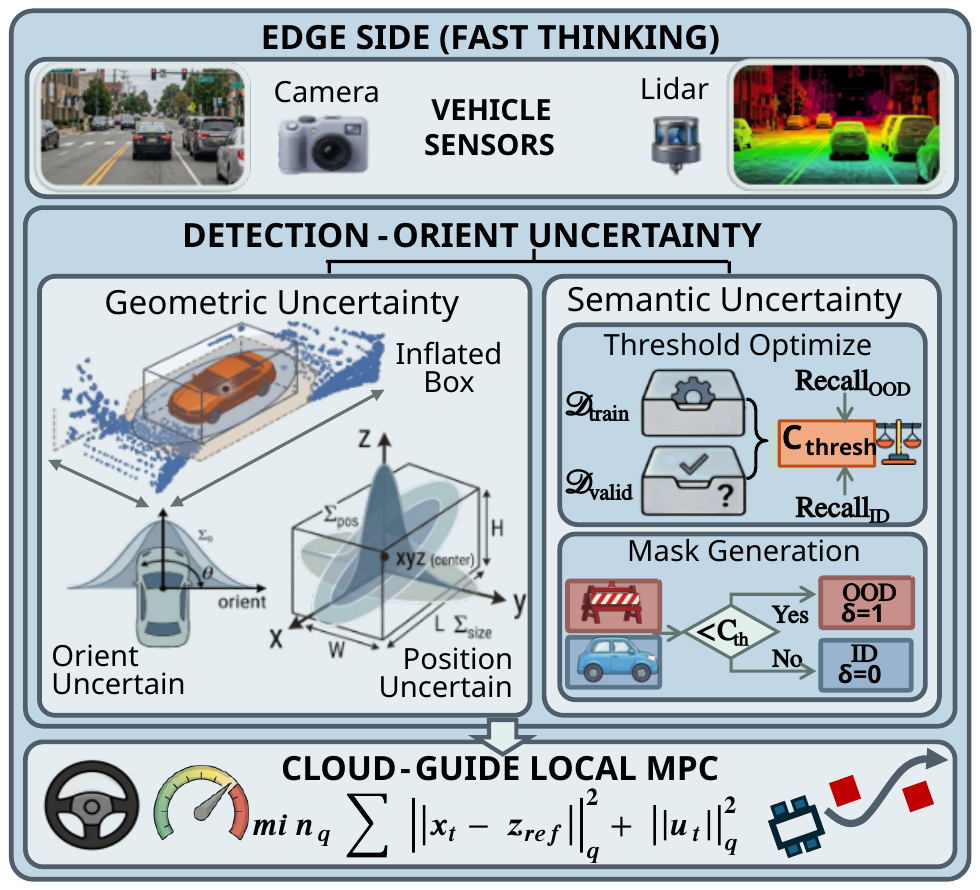}
    \vspace{-3mm}
    \caption{Edge-side uncertainty-aware perception and onboard planning.}
    \label{figure:edge}
    \vspace{-4mm}
\end{figure}

This decomposition provides a structured interface linking perception, planning, and cloud collaboration. Semantic uncertainty identifies candidate ambiguities that may require additional reasoning or external information, whereas geometric uncertainty governs how uncertainty propagates into planning and affects trajectory quality. Accordingly, semantic uncertainty is modeled as a confidence-based screening signal in 2D perception, while geometric uncertainty is represented as a probabilistic 3D object-detection output. The overall edge-side system architecture is illustrated in Fig.~\ref{figure:edge}.

\subsection{Semantic Uncertainty for Unknown Object Detection}
\label{subsec:2d_semantic_ood}

In open-world driving scenarios, local perception models may encounter objects that are rare, unseen, or outside the training distribution. Such cases introduce ambiguity in semantic recognition, as the detector may be unable to confidently determine object identity. In our framework, this ambiguity is modeled as \emph{semantic uncertainty}, which serves as a \emph{decision-triggering signal} indicating that local perception may be unreliable and additional reasoning could be beneficial.

We adopt a simple yet effective confidence-based approximation to operationalize semantic uncertainty in practice. Given an input image at time step $t$, the local 2D object detector produces a set of bounding boxes with associated confidence scores $\{c_{q,t}\}$. Following confidence-based OOD detection principles, low-confidence predictions are treated as indicators of elevated semantic uncertainty. Specifically, we introduce a confidence threshold $c_{\mathrm{th}}$ to distinguish reliable in-distribution detections from potentially out-of-distribution objects. If the confidence score $c_{q,t}$ of a bounding box falls below $c_{\mathrm{th}}$, the corresponding object is marked as semantically uncertain, resulting in a binary screening signal $\delta_{q,t}^{\mathrm{sem}} \in \{0,1\}$ for each detected object $q$.

To determine $c_{\mathrm{th}}$ in a principled and data-driven manner, following confidence-based OOD detection and selective prediction with rejection~\cite{hendrycks2017baseline,geifman2017selective,bendale2016towards}, we adopt the object detection confidence thresholding (ODCT) algorithm. The basic idea is to use the detector confidence as a post-hoc reliability score: high-confidence detections are treated as reliable in-distribution predictions, while low-confidence detections are screened as semantically uncertain and potentially unknown objects. The dataset is partitioned into a training set $\mathcal{D}_{\mathrm{train}} = \{ (\mathbf{x}_i, y_i) \}_{i=1}^{N_{\mathrm{train}}}$ and a validation set $\mathcal{D}_{\mathrm{val}} = \{ (\mathbf{x}_j, y_j) \}_{j=1}^{N_{\mathrm{val}}}$, where $\mathcal{Y}_{\mathrm{train}}$ denotes the set of known semantic classes. The optimal threshold is obtained by solving
\begin{align}
    c_{\mathrm{th}}^\star = \operatorname*{arg\,max}_{c_{\mathrm{th}}\in[0,1]} \ G(c_{\mathrm{th}}),
\end{align}
where $G(c_{\mathrm{th}})$ is defined in \eqref{eq:recall}.
\begin{figure*}[!b]
\normalsize
\hrulefill
\begin{equation}
G(c_{\mathrm{th}}) =
\underbrace{\frac{\sum_{j=1}^{N_{\mathrm{val}}} \mathbb{I}(\hat{y}_j \notin \mathcal{Y}_{\mathrm{train}})\cdot \mathbb{I}(c_j < c_{\mathrm{th}})}{\sum_{j=1}^{N_{\mathrm{val}}} \mathbb{I}(\hat{y}_j \notin \mathcal{Y}_{\mathrm{train}})}}_{\text{recall of OOD detections}}+\underbrace{\frac{\sum_{j=1}^{N_{\mathrm{val}}} \mathbb{I}(\hat{y}_j \in \mathcal{Y}_{\mathrm{train}})\cdot \mathbb{I}(c_j > c_{\mathrm{th}})}{\sum_{j=1}^{N_{\mathrm{val}}} \mathbb{I}(\hat{y}_j \in \mathcal{Y}_{\mathrm{train}})}}_{\text{recall of in-distribution detections}}.
\label{eq:recall}
\end{equation}
\vspace{-0.1in}
\end{figure*}

The objective $G(c_{\mathrm{th}})$ balances the recall of out-of-distribution and in-distribution detections, enabling reliable identification of semantically uncertain objects while suppressing excessive false positives. Importantly, semantic uncertainty does not directly affect planning; instead, it acts as an upstream cue that flags potentially unreliable perception. This signal is subsequently used to evaluate whether resolving such uncertainty is likely to improve downstream planning decisions, thereby supporting opportunity-driven cloud collaboration.

\subsection{Geometric Uncertainty for Bbox Estimation}
\label{subsec:geometric_uncertainty}

In contrast to semantic uncertainty, geometric uncertainty directly affects motion planning by altering obstacle configurations and collision-risk assessment. Even small errors in object position or spatial extent may influence collision checking, feasible trajectory construction, and the final planning result. Therefore, geometric uncertainty should be explicitly modeled and propagated into the planning process, since it determines how uncertain obstacle geometry reshapes the feasible set of the planner and potentially alters the optimal trajectory.

To capture such uncertainty, we formulate 3D bounding box estimation as a probabilistic regression problem. For each detected object $q$, the bounding box is represented by its center position and spatial dimensions. Instead of producing a deterministic box, the detector predicts a Gaussian distribution over the box parameters:

\begin{equation}
\mathbf{z}_q
\triangleq
[x_q, y_q, z_q, l_q, w_q, h_q]^\top 
\sim
\mathcal{N}\!\left(
\boldsymbol{\mu}_q,\boldsymbol{\Sigma}_q
\right)
.
\label{eq:geometric_uncertainty_bbox}
\end{equation}
where
$\boldsymbol{\Sigma}_q\triangleq
\operatorname{diag}\!\left(
\sigma_{q,x}^{2},
\sigma_{q,y}^{2},
\sigma_{q,z}^{2},
\sigma_{q,l}^{2},
\sigma_{q,w}^{2},
\sigma_{q,h}^{2}
\right)$.
Here, $(x_q,y_q,z_q)$ denotes the box center, while $(l_q,w_q,h_q)$ denotes the box dimensions. The vector $\boldsymbol{\mu}_q$ is the predicted mean of the bounding box parameters, and $\boldsymbol{\Sigma}_q$ encodes their geometric uncertainty. For computational efficiency, we adopt an axis-aligned covariance structure, where each box parameter is assigned an independent variance.

The predicted geometric uncertainty is propagated into planning by inflating each obstacle according to planar position uncertainty:
\begin{subequations}
\label{eq:uncertainty_inflated_obstacle}
\begin{align}
\chi_q
&=
\max\{|\Delta p_{q,x}|,|\Delta p_{q,y}|\},\\
r_q^{\mathrm{base}}(\eta)
&=
F_{\chi_q}^{-1}(\eta),\\
\mathcal{O}_q^{\mathrm{infl}}(\eta)
&=
\mathcal{O}_q(\boldsymbol{\mu}_q)
\oplus
\rev{\mathbb{B}_\infty}\!\left(r_q^{\mathrm{base}}(\eta)\right),\\
\Pr\!\left(
\mathbf{p}_q\in\mathcal{O}_q^{\mathrm{infl}}(\eta)
\right)
&\geq \eta .
\end{align}
\end{subequations}
Here, $F_{\chi_q}^{-1}(\cdot)$ is the quantile function of the maximum planar perturbation $\chi_q$, $\oplus$ denotes Minkowski addition, and \rev{$\mathbb{B}_\infty(r)=\{\delta:\|\delta\|_\infty\le r\}$ is the planar $L_\infty$ ball. Thus, larger uncertainty yields a larger safety envelope}.

To effectively learn the uncertainty parameters, we augment the detector with a Gaussian uncertainty head that jointly predicts the bounding box mean $\boldsymbol{\mu}_q$ and the log-variance vector $\mathbf{s}_q=\log\boldsymbol{\sigma}_q^{2}$. The uncertainty head is trained using the Gaussian negative log-likelihood: 
\begin{align}
\mathcal{L}_{\mathrm{reg}}(\mathbf{z}_q,\boldsymbol{\mu}_q,\mathbf{s}_q)
&=
\sum_{d=1}^{6}
\frac{\exp(-s_{q,d})
(z_{q,d}-\mu_{q,d})^2
+
s_{q,d}}{2}.
\nonumber
\end{align}
The element-wise squared error $(z_{q,d}-\mu_{q,d})^2$ is weighted by the predicted inverse variance $\exp(-s_{q,d})$ for each bounding-box parameter $d$. The log-variance vector $\mathbf{s}_q$ allows the detector to assign uncertainty to ambiguous or noisy observations while penalizing unnecessarily large variance. For numerical stability, $\mathbf{s}_q$ is clamped within a bounded range during training. 
As such, the predicted uncertainty is further used to calibrate detection confidence:
\begin{align}
c_q^{\mathrm{cls}}
&=
\operatorname{sigmoid}(\ell_q^{\mathrm{raw}})
\cdot
g(\mathbf{s}_q),
\nonumber
\end{align}
where $\ell_q^{\mathrm{raw}}$ is the raw classification logit, $\operatorname{sigmoid}(\ell_q^{\mathrm{raw}})$ is the confidence, and \rev{$g(\mathbf{s}_q)=\operatorname{sigmoid}(f_g(\mathbf{s}_q))\in(0,1)$ is a learned uncertainty-conditioned gating function, where $f_g(\cdot)$ is implemented by a lightweight MLP over the predicted log-variance features}. When the geometric prediction is uncertain, this gating term reduces the detection confidence.

Overall, the learned Gaussian parameters $(\boldsymbol{\mu}_q,\boldsymbol{\Sigma}_q)$ provide a probabilistic representation of obstacle geometry that can be directly integrated into planning modules. This allows geometric uncertainty to be explicitly propagated into trajectory optimization and supports the subsequent task-oriented value estimation framework, where the impact of uncertainty is quantified according to its influence on decision-making. An implementation of the 3D bounding-box uncertainty module is publicly available at \url{https://github.com/cjychenjiayi/3D_Bbox_uncertainty}.

\section{Expected Planning Gain
\\Under Perception Uncertainty}
\label{sec:value_modeling}

In Section~\ref{sec:uncertainty_modeling}, perception uncertainty is decomposed into semantic and geometric components. However, its practical importance depends on whether resolving it changes the downstream planning decision. We therefore evaluate uncertainty through EPG, which estimates the expected normalized MPC cost reduction over plausible scene-uncertainty realizations induced by the current observation. This provides a task-specific approximation of the value of additional information without maintaining a full belief-space planner.

Following this principle, we propagate perception uncertainty through the planning pipeline. Geometric uncertainty is first converted into a probabilistic obstacle representation (Section~\ref{subsec:probabilistic_obstacle}), which is then incorporated into a dynamics-aware MPC optimization problem to define planning cost under each scene realization (Section~\ref{subsec:mpc_planning}). Finally, a budgeted Monte Carlo procedure samples plausible scene realizations and evaluates their corresponding planning costs, producing a scalar EPG estimate that captures the expected impact of uncertainty resolution (Section~\ref{subsec:mc_value}).

\begin{figure}[!t]
    \centering
    \includegraphics[width=0.92\columnwidth]{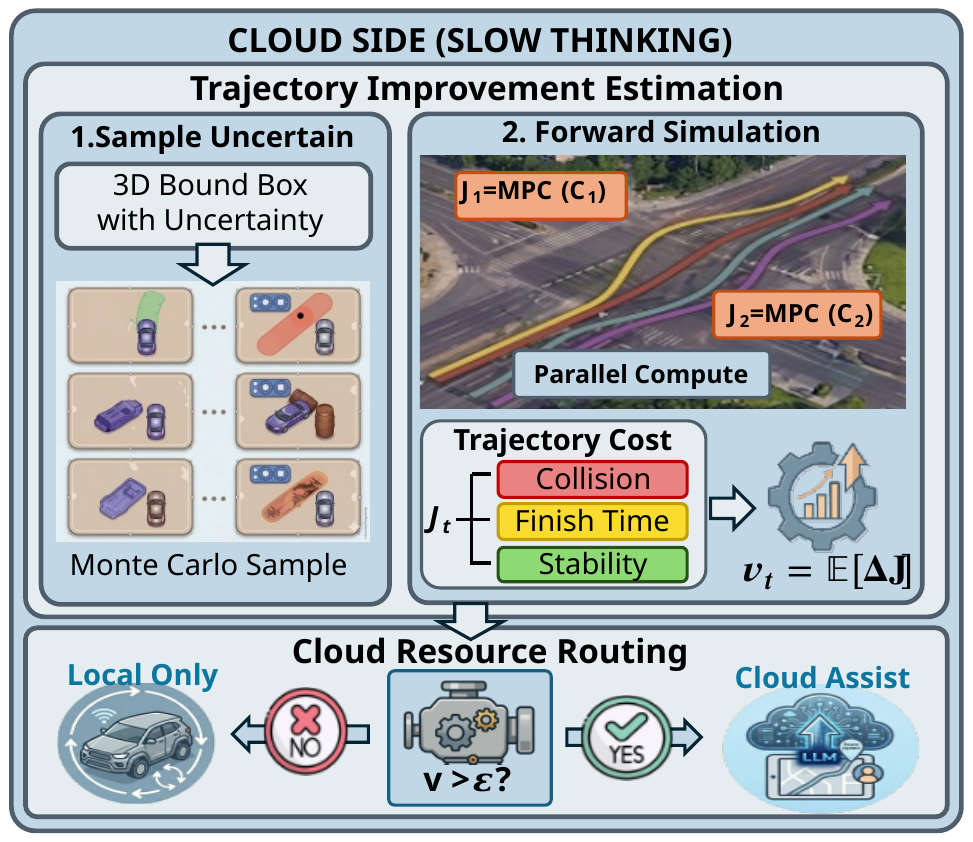}
    \vspace{-2mm}
    \caption{Cloud-side slow-reasoning for planning guidance.}
    \label{figure:cloud}
    \vspace{-4mm}
\end{figure}

\subsection{Safety-Constrained Probabilistic Obstacle Modeling}
\label{subsec:probabilistic_obstacle}

The goal of this section is to propagate geometric uncertainty into motion planning and quantify its effect on obstacle constraints, so that its contribution to downstream planning gain can be evaluated. While semantic uncertainty mainly serves as a screening signal for cloud assistance, motion planning is directly affected by geometric uncertainty, which determines obstacle configurations and collision risk. Therefore, we construct a probabilistic obstacle representation that explicitly links perception uncertainty to planning constraints.

Let $\mathcal{O}_q^0$ denote the nominal occupied region induced by the 3D bounding box of a detected object $q$. Based on the geometric uncertainty estimated in Section~\ref{sec:uncertainty_modeling}, the planning-relevant uncertainty of the obstacle is modeled through a randomly translated obstacle region:
\begin{subequations}
\label{eq:probabilistic_obstacle_geometry}
\begin{align}
\mathcal{O}_q^{\mathrm{true}}
&=
\mathcal{O}_q^0
\oplus
\Delta \mathbf{p}_q, \ 
\mathrm{with} \ 
\Delta \mathbf{p}_q \sim
\mathcal{N}
\left(
\mathbf{0},
\mathbf{\Sigma}_{q}
\right).
\end{align}
\end{subequations}
Here, 
$
\mathbf{\Sigma}_{q}=
\operatorname{diag}
\left(
\sigma_{q,x}^{2},
\sigma_{q,y}^{2},
\sigma_{q,z}^{2}
\right)$, $\Delta \mathbf{p}_q$ denotes uncertain object-center displacement, and $\oplus$ translates the nominal box by this displacement. Thus, $\mathcal{O}_q^{\mathrm{true}}$ represents a random obstacle region induced by perception uncertainty.

In general, safe navigation under this uncertainty can be formulated as a chance-constrained collision avoidance requirement~\cite{blackmore2011chance,dawson2020provably}. Specifically, for an ego-vehicle trajectory $\tau$, the probability of remaining collision-free with respect to object $q$ should be no smaller than a prescribed safety confidence level $\eta_{\mathrm{safe}}$:
\begin{equation}
\Pr\!\left(
\inf_t
\operatorname{dist}
\bigl(
\tau(t),
\mathcal{O}_q^{\mathrm{true}}
\bigr)
>
0
\right)
\geq
\eta_{\mathrm{safe}} .
\label{eq:chance_constrained_safety}
\end{equation}
This constraint captures the desired trajectory-level safety requirement. However, directly enforcing it in real time is computationally challenging, since it requires probabilistic collision checking over random obstacle geometry. As a tractable compromise, we convert this chance-constrained safety requirement into a deterministic inflated-obstacle constraint in the following.

To establish a tractable approximation, we convert the probabilistic safety requirement into a deterministic inflated-obstacle constraint. In practice, since collision checking is mainly determined by horizontal displacement, we define the inflation radius as the $\eta_{\mathrm{safe}}$-quantile of the worst-case planar perturbation~\cite{du2011probabilistic,dawson2020provably}:
\begin{equation}
\label{eq:inflated_obstacle_mixture_a}
\begin{aligned}
r_q^{\mathrm{base}}
&\triangleq
F_{\chi_q}^{-1}\!\left(\eta_{\mathrm{safe}}\right)
=
\inf\!\left\{
r\ge 0:
F_{\chi_q}(r)\ge \eta_{\mathrm{safe}}
\right\}.
\end{aligned}
\end{equation}
Here, $\chi_q
=
\max
\left\{
|\Delta p_{q,x}|,
|\Delta p_{q,y}|
\right\}$
\rev{measures} the maximum horizontal perturbation relevant to planner collision checking,
and $F_{\chi_q}^{-1}(\cdot)$ denotes the quantile function of $\chi_q$.
Thus, $r_q^{\mathrm{base}}$ corresponds to the prescribed safety-level quantile of the planar uncertainty. \footnote{In implementation, the distribution of $\chi_q$ can be obtained either analytically from the predicted Gaussian perturbation when an independent diagonal covariance is assumed, or numerically by Monte Carlo sampling. We use the resulting empirical cumulative distribution to evaluate the safety-level quantile.}

Leveraging the inflation radius, the nominal obstacle is converted into a safety envelope. When the planar uncertainty is multi-modal, the same inflated-obstacle form is retained, while the radius is determined by the corresponding mixture-distribution quantile:
\begin{equation}
\label{eq:inflated_obstacle_mixture_b}
\mathcal{O}_q^{\mathrm{infl}}
\triangleq
\mathcal{O}_q^0
\oplus
\rev{\mathbb{B}_\infty}\!\left(r_q^{\mathrm{base}}\right).
\end{equation}
The probabilistic obstacle representation establishes a direct bridge between perception uncertainty and motion planning constraints. By translating uncertain obstacle geometry into safety-constrained inflated obstacle sets, geometric uncertainty reshapes the feasible trajectory set, thereby providing the planning constraints for subsequent EPG estimation.

\subsection{Planning-Gain Evaluation via MPC}
\label{subsec:mpc_planning}

Given a deterministic scene configuration $\mathcal{C}_t$, the planner evaluates its decision-level effect by solving an MPC problem. Let $\mathbf{x}_{k|t}$ and $\mathbf{u}_{k|t}$ denote the predicted state and control at step $k$ from time $t$, respectively. The time-indexed obstacle configuration is defined as
\begin{equation*}
\mathcal{C}_t=
\left\{
(\mathcal{O}^{0}_{q,k|t},r_{q,k|t})
\;:\;
q=1,\ldots,M,
\ k=0,\ldots,N
\right\},
\end{equation*}
which covers both static and dynamic obstacles. The MPC planner minimizes tracking error, control effort, and control smoothness subject to the dynamics, state/control limits, and time-indexed collision-avoidance constraints:
\begin{align}
J_t(\mathcal{C}_t)
=&
\min_{\{\mathbf{x}_{k|t},\mathbf{u}_{k|t}\}}
\nonumber
\sum_{k=0}^{N}
\left\|
\mathbf{x}_{k|t}
-
\mathbf{x}_{k|t}^{\mathrm{ref}}
\right\|_{\mathbf{Q}}^{2}
+
\sum_{k=0}^{N-1}
\left\|
\mathbf{u}_{k|t}
\right\|_{\mathbf{R}}^{2}
\nonumber\\
&
\quad\quad\quad\quad
+
\sum_{k=1}^{N-1}
\left\|
\mathbf{u}_{k|t}
-
\mathbf{u}_{k-1|t}
\right\|_{\mathbf{S}}^{2}
\nonumber
\\
\mathrm{s.t.}\quad
&
\mathbf{x}_{0|t}=\mathbf{x}_t,
\nonumber\\
&
\mathbf{x}_{k+1|t}
=
f(\mathbf{x}_{k|t},\mathbf{u}_{k|t}),
\quad
k=0,\ldots,N-1,
\nonumber\\
&
\mathbf{x}_{k|t}\in\mathcal{X},
\quad
k=0,\ldots,N,
\nonumber\\
&
\mathbf{u}_{k|t}\in\mathcal{U},
\quad
k=0,\ldots,N-1,
\nonumber\\
&
g_{\mathrm{coll}}
\left(
\mathbf{x}_{k|t},
\mathcal{O}^{0}_{q,k|t},
r_{q,k|t}
\right)
\leq 0,
\nonumber\\
&
q=1,\ldots,M,
\ k=0,\ldots,N.
\label{eq:mpc_planning_cost}
\end{align}
The resulting optimal predicted trajectory is
\begin{equation}
\tau_t^{\star}(\mathcal{C}_t)
=
\{\mathbf{x}^{\star}_{k|t}(\mathcal{C}_t)\}_{k=0}^{N}.
\end{equation}
For two scene configurations $\mathcal{C}_t^{1}$ and $\mathcal{C}_t^{2}$, the normalized planning gain is defined as
\begin{equation}
\Delta_{\mathrm{pg}}(\mathcal{C}_t^{1},\mathcal{C}_t^{2})
=
\frac{
J_t(\mathcal{C}_t^{1})-J_t(\mathcal{C}_t^{2})
}{
J_t(\mathcal{C}_t^{1})
}.
\label{eq:planning_gain_pairwise}
\end{equation}
A negligible $\Delta_{\mathrm{pg}}$ means that resolving the uncertainty does not materially change the planner, whereas a large value indicates decision-relevant uncertainty.

\begin{figure*}[!t]
    \centering
    \scalebox{1.1}[1.00]{%
        \includegraphics[width=0.85\textwidth]{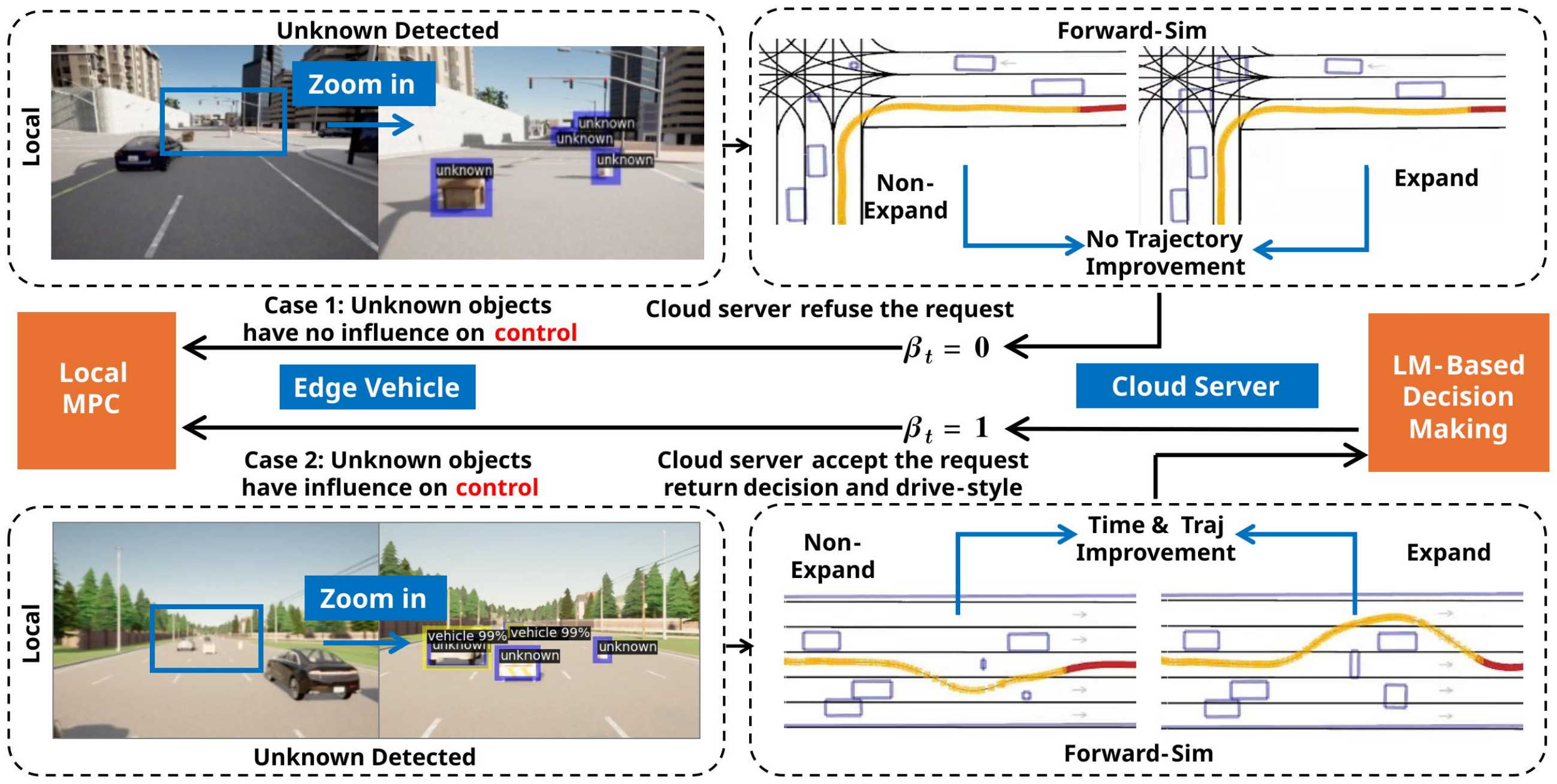}
    }
    \vspace{-2mm}
    \caption{Cloud-assisted forward simulation for decision evaluation.}
    \label{figure:cfs}
    \vspace{-2mm}
\end{figure*}

\subsection{Budget-Aware Monte Carlo Estimation}
\label{subsec:mc_value}

This module estimates whether resolving uncertainty would improve downstream planning. Instead of relying on perception-level uncertainty alone, SIGMA propagates plausible uncertainty realizations through the MPC planner and evaluates their planning-cost impact, as illustrated in Fig.~\ref{figure:cfs}.

Let $\mathcal{C}_t^{\mathrm{base}}$ denote the conservative obstacle configuration constructed from local perception, and let $\boldsymbol{\xi}_t=\{\Delta \mathbf{p}_{q,t}\}_{q=1}^{M}$ denote a plausible realization of the current scene uncertainty. The corresponding uncertainty-resolved scene configuration is denoted by $\mathcal{C}_t(\boldsymbol{\xi}_t)$. We define
\begin{subequations}
\label{eq:epg_mc_definition}
\begin{align}
\delta J_t(\boldsymbol{\xi}_t)
&=
\frac{
J_t(\mathcal{C}_t^{\mathrm{base}})
-
J_t(\mathcal{C}_t(\boldsymbol{\xi}_t))
}{
J_t(\mathcal{C}_t^{\mathrm{base}})
},\\
\mathrm{EPG}_t
&=
\mathbb{E}_{\boldsymbol{\xi}_t\sim p_t(\boldsymbol{\xi}\mid \mathbf{o}_t)}
\left[
\delta J_t(\boldsymbol{\xi}_t)
\right].
\end{align}
\end{subequations}
Here, $\delta J_t(\boldsymbol{\xi}_t)$ is the normalized MPC cost reduction under a plausible uncertainty-resolved scene realization, and the expectation is taken over the uncertainty distribution induced by the current observation.

Since this expectation is generally intractable, we estimate it using importance-sampling-based Monte Carlo rollout~\cite{robert2004monte,kahn1953methods}. Assuming conditional independence among object-level perturbations,
\begin{equation}
p_t(\boldsymbol{\xi}\mid \mathbf{o}_t)
=
\prod_{q=1}^{M}
p_{q,t}(\Delta \mathbf{p}_{q,t}\mid \mathbf{o}_t).
\end{equation}
Given samples $\boldsymbol{\xi}_t^{(i)}\sim\phi_t(\boldsymbol{\xi})$, the weighted estimator is
\begin{subequations}
\label{eq:epg_importance_sampling}
\begin{align}
w_i
&=
\frac{
p_t(\boldsymbol{\xi}_t^{(i)}\mid \mathbf{o}_t)
}{
\phi_t(\boldsymbol{\xi}_t^{(i)})
},\\
\widehat{\mathrm{EPG}}_t
&=
\frac{
\sum_{i=1}^{K}
w_i\delta J_t(\boldsymbol{\xi}_t^{(i)})
}{
\sum_{i=1}^{K}w_i
}.
\end{align}
\end{subequations}
The estimate $\widehat{\mathrm{EPG}}_t$ is used as the planning-level criterion for selective cloud invocation: uncertainties with negligible estimated planning gain are filtered locally, while decision-relevant cases are preserved for cloud-assisted reasoning.

\section{Fast--Slow Collaborative Decision and Planning-Gain-Aware Cloud Resource Allocation}

Autonomous driving naturally exhibits a fast--slow decision modeling structure. Most driving scenarios can be handled effectively by fast onboard processing, which provides real-time perception, planning, and control. However, a small subset of complex or ambiguous situations requires deeper reasoning and more reliable perception, which can be enabled by cloud-side large models. The key challenge is therefore to determine \emph{when} cloud assistance is expected to provide meaningful decision-level benefit under uncertainty and limited system resources.

In our framework, the EPG defined in Section~\ref{sec:value_modeling} serves as the decision-level criterion for fast--slow collaboration. In online operation, EPG is not directly observable and is replaced by its simulation-in-the-loop Monte Carlo estimate, denoted by $\widehat{\mathrm{EPG}}$. Specifically, $\widehat{\mathrm{EPG}}$ governs three aspects of the system: (i) cloud activation, namely whether the current scenario should be escalated to the cloud; (ii) decision refinement, namely how cloud intelligence improves the driving decision; and (iii) resource allocation, namely how limited cloud computation is distributed across multiple requests. In this way, cloud collaboration is driven by estimated planning impact rather than perception uncertainty or heuristic triggers.

Cloud assistance is activated only when $\widehat{\mathrm{EPG}}$ is sufficiently large, indicating that resolving the current uncertainty is estimated to improve the downstream planning outcome. Otherwise, the scenario is handled entirely onboard, avoiding unnecessary latency and computation. Once activated, the cloud refines perception and performs high-level semantic reasoning to generate guidance, which is then incorporated into the local planner. When multiple vehicles simultaneously request assistance, the same planning-gain principle is further used to prioritize cloud resources across competing requests.

\subsection{LVM-Assisted Semantic Decision Generation}
\label{subsec:lvm_decision}

The cloud-side decision module is invoked only when the onboard system identifies decision-relevant uncertainty with a sufficiently large EPG. In such cases, the cloud module performs enhanced perception and semantic reasoning to provide high-level planning guidance that may not be reliably obtained from local processing alone.

We adopt a two-stage paradigm consisting of perception refinement and semantic decision generation. Given the input image $\mathbf{x}_t$, a high-accuracy cloud perception model first produces refined object detections, which are then organized into a structured semantic scene representation:
\label{eq:cloud_semantic_representation}
\begin{align}
\mathcal{B}_t^\diamond=
g_{\mathrm{cloud}}(\mathbf{x}_t), \
\mathcal{Z}_t=
\Phi
\left(
\mathcal{B}_t^\diamond,
\mathcal{L}_t,
\mathcal{R}_t
\right).
\end{align}
Here, $\mathcal{B}_t^\diamond$ denotes the refined cloud-side detection results, while $\mathcal{Z}_t$ represents the structured semantic scene representation. The function $\Phi(\cdot)$ integrates object categories, spatial relationships, lane-level structure $\mathcal{L}_t$, and drivable-region information $\mathcal{R}_t$ into a decision-oriented scene description.

\begin{figure}[!t]
    \centering
    \includegraphics[width=0.95\columnwidth]{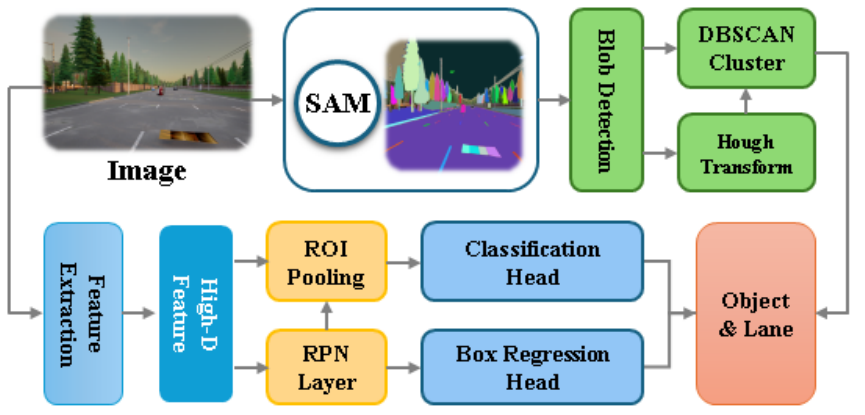}
    \vspace{-1mm}
    \caption{Cloud-based LVM perception pipeline with SAM-assisted.}
    \label{figure:sam}
    \vspace{-3mm}
\end{figure}

To improve robustness in complex scenarios, we incorporate segmentation-based verification, such as SAM, as illustrated in Fig.~\ref{figure:sam}. The segmentation results are adopted to refine geometric structure, suppress unreliable detections, and extract drivable regions and lane geometry. The lane structure is obtained through geometric filtering and curve fitting:
\label{eq:segmentation_lane_refinement}
\begin{align}
\mathcal{S}_t
=
h_{\mathrm{seg}}(\mathbf{x}_t,\mathcal{B}_t^\diamond), \
\mathcal{L}_t
=
\left\{
y=a_jx+b_j
\right\}_{j=1}^{N_{\mathrm{lane}}}.
\end{align}
Here, $\mathcal{S}_t$ denotes the segmentation-based verification result, and $\mathcal{L}_t$ denotes the reconstructed lane structure. Each lane candidate is represented by a fitted line parameterized by $(a_j,b_j)$, where $N_{\mathrm{lane}}$ is the number of detected lane structures. The resulting lane and drivable-region information complements the refined object detections and improves the reliability of the semantic scene representation.

\begin{figure}[!t]
    \centering
    \scalebox{1.02}[1.00]{%
        \includegraphics[width=0.9\columnwidth]{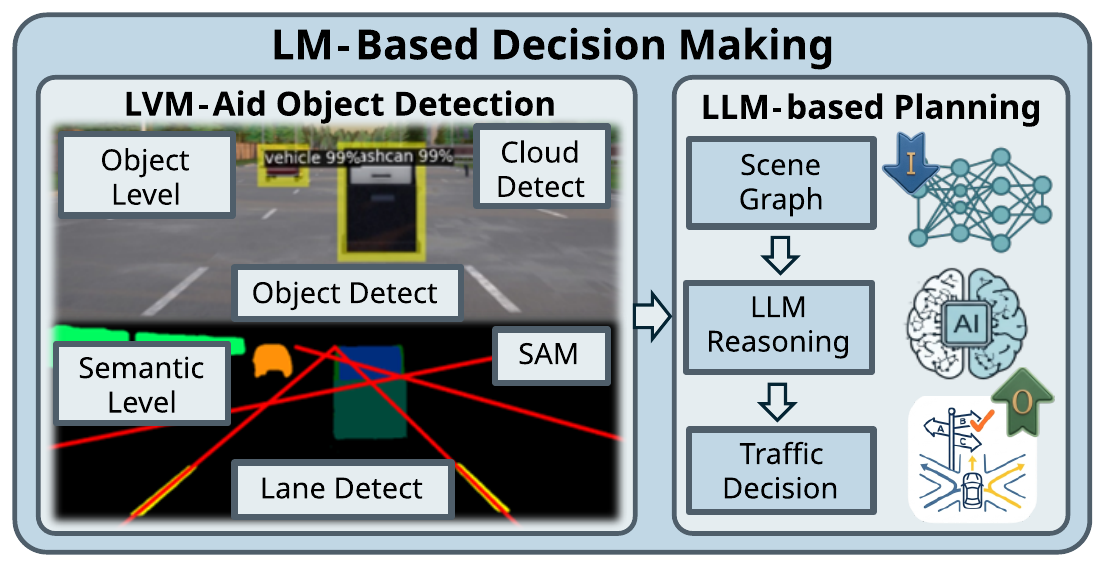}
    }
    \vspace{-2mm}
    \caption{Cloud-based large model decision-making pipeline.}
    \label{figure:lm_decision}
    \vspace{-3mm}
\end{figure}

Based on $\mathcal{Z}_t$, the large vision--language model performs semantic reasoning over obstacles, lane structures, and drivable regions to generate high-level traffic decisions, as shown in Fig.~\ref{figure:lm_decision}. Specifically, following the agentic fast--slow planning strategy in~\cite{chen2026bridging}, the cloud-side model first produces a serialized traffic-decision sequence that describes the intended maneuver and its contextual rationale. This sequence is then parsed into maneuver-level guidance and converted into a reference trajectory for the local MPC planner:
\begin{subequations}
\label{eq:lvm_decision_reference}
\begin{equation}
\gamma_t
=
\Psi_{\mathrm{LVM}}(\mathcal{Z}_t)
\in
\{\mathrm{keep}, \mathrm{left}, \mathrm{right}\},
\end{equation}
\begin{equation}
\mathcal{W}_t^\diamond
=
\mathcal{T}(\gamma_t, \mathcal{Z}_t).
\end{equation}
\end{subequations}
Here, $\Psi_{\mathrm{LVM}}(\cdot)$ denotes the semantic reasoning function of the large vision--language model, which follows the agentic fast--slow planning paradigm to produce the serialized traffic-decision sequence. The parsed maneuver $\gamma_t$ specifies whether the ego vehicle should keep its lane, change to the left lane, or change to the right lane. The trajectory generation function $\mathcal{T}(\cdot)$ converts the maneuver decision and semantic scene representation into a reference trajectory $\mathcal{W}_t^\diamond$, which is provided to the local MPC module. In this way, the cloud module provides semantically informed and long-horizon planning guidance, while the onboard MPC module retains real-time control authority and enforces dynamic feasibility and safety.

\subsection{EPG- and Deadline-Aware Cloud-Guided MPC}
\label{subsec:cloud_guided_mpc}

When multiple vehicles request cloud assistance, the cloud-side module should allocate its limited computation to requests whose uncertainty is both decision-relevant and actionable before the corresponding driving deadline. In this work, we adopt a lightweight EPG- and deadline-aware rule as an initial resource allocation mechanism, rather than solving a full-cloud scheduling optimization problem.

For each request $i$, let $\widehat{\mathrm{EPG}}_i$ denote the simulation-in-the-loop estimate of the expected planning gain obtained in Section~V-C. This quantity estimates the normalized MPC planning-cost reduction that could be achieved by resolving the current uncertainty. Let $t_i^{\mathrm{obs}}$ denote the predicted arrival time at the first decision-critical obstacle. The remaining decision slack and the predicted cloud response time are defined as
\begin{equation}
s_i(t)
=
t_i^{\mathrm{obs}}-t,
\rev{\quad}
\widehat{\Delta}_i^{\mathrm{inf}}
=
\widehat{\Delta}_i^{\mathrm{comm}}
+
\widehat{\Delta}_i^{\mathrm{cloud}} .
\label{eq:cloud_latency}
\end{equation}
Here, $s_i(t)$ is the remaining time before the ego vehicle reaches the first obstacle that may require a planning decision, and $\widehat{\Delta}_i^{\mathrm{inf}}$ is the predicted cloud response time, including communication and cloud-side inference. Cloud assistance is useful only if the response is predicted to return within this slack. \rev{Following the benefit-factor formulation in Section~\ref{sec:system_overview}, we define the request-level benefit factor as}
\begin{equation}
\rev{\tilde{g}_i(t)}
=
\widehat{\mathrm{EPG}}_i
\rev{-\lambda \kappa_i^{\mathrm{lm}}
-\rho D(\widehat{\Delta}_i^{\mathrm{inf}})} .
\label{eq:request_benefit_factor}
\end{equation}
\rev{where $\kappa_i^{\mathrm{lm}}$ denotes the invocation cost of request $i$.}
Therefore, the cloud request decision is defined by a \rev{benefit-factor} threshold and a deadline feasibility condition:
\begin{equation}
\alpha_i(t)
=
\mathbb{I}
\left(
\rev{\tilde{g}_i(t)>\theta_g}
\right)
\mathbb{I}
\left(
s_i(t)>\widehat{\Delta}_i^{\mathrm{inf}}
\right).
\label{eq:cloud_request_decision}
\end{equation}
Here, $\alpha_i(t)=1$ means that request $i$ is eligible for cloud service, while $\alpha_i(t)=0$ means that the request is handled locally. The first condition filters out \rev{requests whose benefit factor is insufficient}, while the second condition ensures that the cloud result is predicted to arrive before the vehicle reaches the decision-critical obstacle.

When the cloud has limited computation capacity, only requests satisfying $\alpha_i(t)=1$ are considered. These candidate requests are ranked by a \rev{benefit-factor-efficiency} score:
\begin{subequations}
\label{eq:cloud_request_priority}
\begin{align}
\mathcal{Q}_{\mathrm{cloud}}(t)
&=
\left\{
i\in\mathcal{Q}(t)
\mid
\alpha_i(t)=1
\right\}, \\
\psi_i(t)
&=
\frac{
\rev{\tilde{g}_i(t)}
}{
\widehat{\Delta}_i^{\mathrm{inf}}+\epsilon
}.
\end{align}
\end{subequations}
Here, $\epsilon>0$ is a small constant introduced to avoid division by zero and improve stability. The cloud serves requests in $\mathcal{Q}_{\mathrm{cloud}}(t)$ according to descending $\psi_i(t)$ until available computation budget is exhausted. This rule favors requests with \rev{large benefit factor} and short predicted response latency.

Importantly, cloud request acceptance does not imply that cloud guidance is immediately available to the local planner. We therefore introduce another binary variable $\beta_i(t)\in\{0,1\}$ to indicate whether valid cloud guidance has been returned before the decision deadline:
\begin{equation}
\beta_i(t)
=
\mathbb{I}
\left(
\alpha_i(t)=1
\right)
\mathbb{I}
\left(
t_i^{\mathrm{ret}}\leq t_i^{\mathrm{ddl}}
\right)
\mathbb{I}
\left(
\rev{\mathrm{Valid}(\mathcal{W}_i^{\diamond})=1}
\right),
\label{eq:cloud_guidance_availability}
\end{equation}
where $t_i^{\mathrm{ret}}$ is the actual cloud-result return time, $t_i^{\mathrm{ddl}}$ is the decision deadline, and $\mathcal{W}_i^{\diamond}$ denotes the cloud-guided reference trajectory. \rev{Here, $\mathrm{Valid}(\mathcal{W}_i^{\diamond})=1$ means that the returned reference is consistent with the current ego state, satisfies drivable-region constraints, and yields a feasible MPC problem. If this check fails, we set $\beta_i(t)=0$, and the planner falls back to the local reference in the reference-switching rule below.} Thus, $\alpha_i(t)$ represents the cloud request decision, whereas $\beta_i(t)$ represents the availability of valid returned cloud guidance.

Once valid cloud guidance is available, the local planner incorporates it by switching the reference path used in the MPC objective:
\begin{equation}
\mathcal{W}_i^{\mathrm{ref}}
=
\begin{cases}
\mathcal{W}_i^{\mathrm{loc}}, & \beta_i(t)=0, \\
\mathcal{W}_i^{\diamond}, & \beta_i(t)=1 .
\end{cases}
\label{eq:reference_switching}
\end{equation}
Here, $\mathcal{W}_i^{\mathrm{loc}}$ is the local reference path, and $\mathcal{W}_i^{\diamond}$ is the cloud-guided reference path. The selected $\mathcal{W}_i^{\mathrm{ref}}$ is then used as the MPC reference input defined in Section~V-B. In this way, the cloud does not directly control the vehicle; it only updates the high-level reference when valid guidance returns before the deadline, while the onboard MPC enforces dynamic feasibility and collision safety in real time.

\section{Experiments}

We evaluate the proposed framework from three perspectives: 1) perception and uncertainty modeling, 2) simulation-in-the-loop planning-gain assessment in static scenarios, and 3) dynamic-scene decision making. All system-level experiments are implemented in Python and ROS on the CARLA platform~\cite{dosovitskiy2017carla}. The ego vehicle is equipped with an RGB-D camera and a $64$-line LiDAR operating at $10$\,Hz. The sensor frequency specifies the perception update rate in simulation, and no additional sensing delay is modeled. The local detection model processes sensory inputs at up to $100$\,FPS and the local MPC module generates control outputs at $20$\,Hz. Also, the safety distance of MPC is set to $d_0=0.3$\,m and $d_{\text{inf}}=3.0$\,m. The prediction horizon is $H=18$ with a time step of $\Delta t = 0.3$\,s. All experiments are conducted on an Ubuntu workstation with two NVIDIA RTX 3090 GPUs.

For comparison, we consider the following schemes: 1) \textbf{Local-only strategy (LOS)}, which exploits local perception and a state-of-the-art MPC (i.e., RDA)~\cite{han2023rda}; 2) \textbf{Periodical collaboration strategy (PCS)}, a baseline collaboration strategy with fixed intervals between consecutive cloud services~\cite{tanwani2020rilaas}; and 3) \textbf{SIGMA (ours)}, the proposed fast--slow collaboration framework based on decision-oriented uncertainty representation, ODCT-based semantic screening, and simulation-in-the-loop EPG estimation. Recent extensions of this work further explore joint query-service optimization~\cite{chen2025joint} and agentic fast--slow planning~\cite{chen2026bridging}. We note that the collaborative planner in~\cite{li2025edge} does not involve large vision models and exhibits similar behavior to LOS~\cite{han2023rda}. We first validate the perception and uncertainty modeling components of the proposed framework, including semantic uncertainty for unknown-object awareness and geometric uncertainty for 3D localization ambiguity. We then evaluate how these uncertainty estimates are exploited by the simulation-in-the-loop planning-gain assessment mechanism in static and dynamic driving scenarios.

\subsection{Experiment 1: Perception and Uncertainty Modeling}

We first validate the perception-level components of the proposed framework from two aspects. 
We evaluate \emph{semantic uncertainty} for unknown-object awareness and cloud-assisted recognition, and then evaluate \emph{geometric uncertainty} for 3D object localization. 
The former verifies whether local perception can reliably identify potentially unknown objects and screen candidate cases for cloud assistance, as described in Section~\ref{subsec:2d_semantic_ood}, while the latter examines whether the predicted geometric uncertainty reflects true localization ambiguity and is therefore suitable for downstream planning.

\begin{figure}[!t]
    \centering
    \includegraphics[width=0.98\columnwidth]{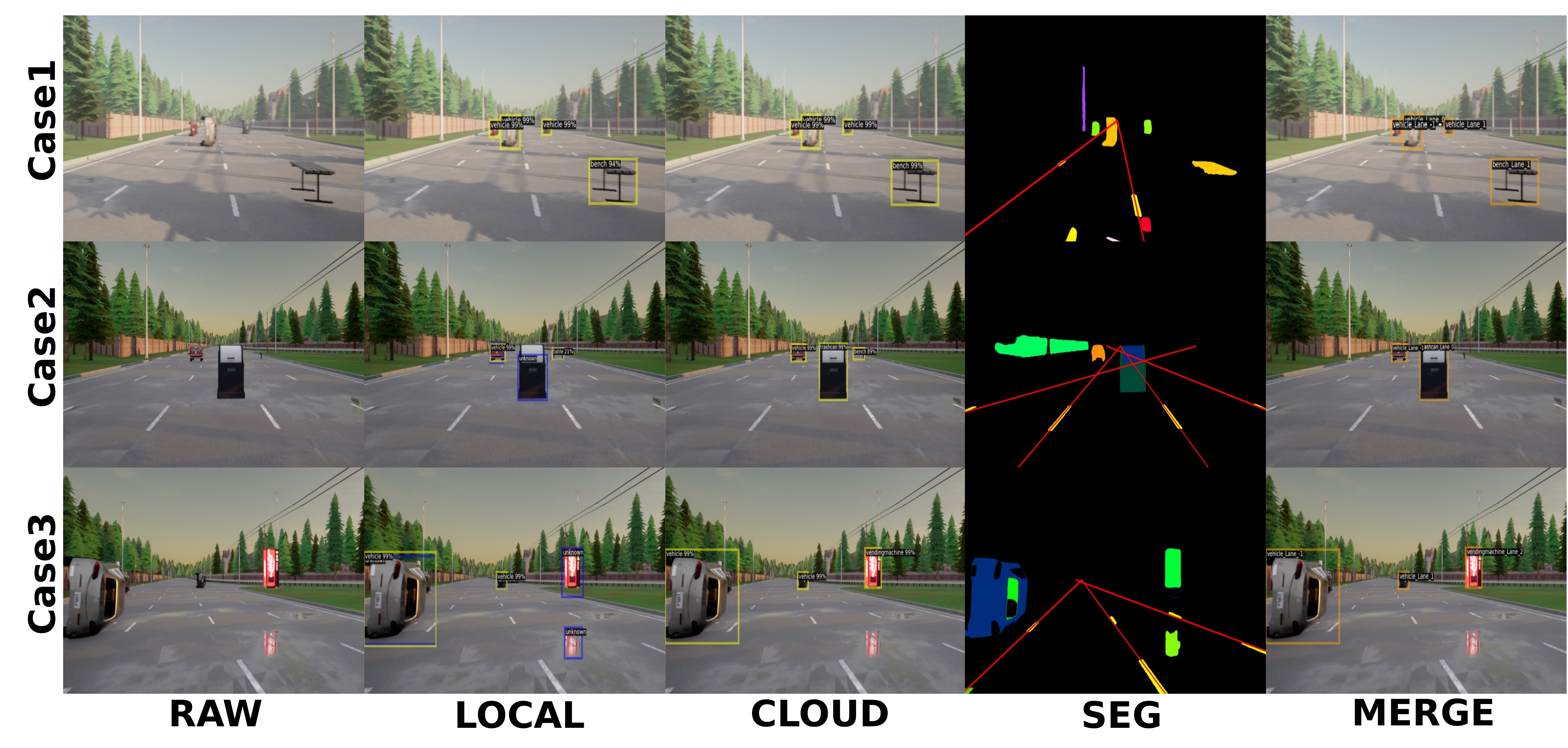}
    \vspace{-1mm}
    \caption{Comparison of local and cloud perception results.}
    \label{fig:semantic_compare}
    \vspace{-3mm}
\end{figure}

\begin{table}[!t]
\centering
\caption{Dataset splits for known- and unknown-object evaluation.}
\label{tab:data_splits}
\setlength{\tabcolsep}{6pt}
\renewcommand{\arraystretch}{1.1}
\begin{tabular}{lcccc}
\toprule
\textbf{Split} & \textbf{\# Samples} & \textbf{Known} & \textbf{Unknown} & \textbf{Total} \\
\midrule
Local\_Train  & 9,647  & 38 & 0 & 38 \\
Local\_Val    & 2,060  & 21 & 5 & 26 \\
Local\_Test   & 3,400  & 19 & 6 & 25 \\
Cloud\_Train  & 15,107 & 49 & 0 & 49 \\
\bottomrule
\end{tabular}
\vspace{-2mm}
\end{table}

\begin{figure}[!t]
    \centering
    \subfloat[Validation set.]{
        \includegraphics[width=0.45\columnwidth]{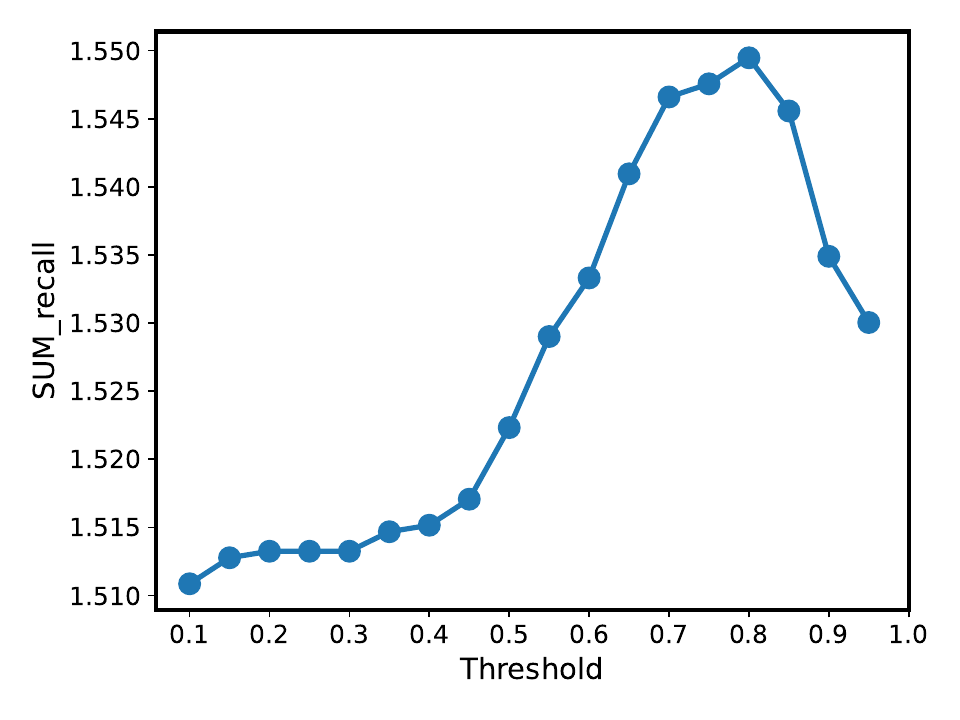}
        \label{fig:base_scena_val}
    }
    \hfil
    \subfloat[Test set.]{
        \includegraphics[width=0.45\columnwidth]{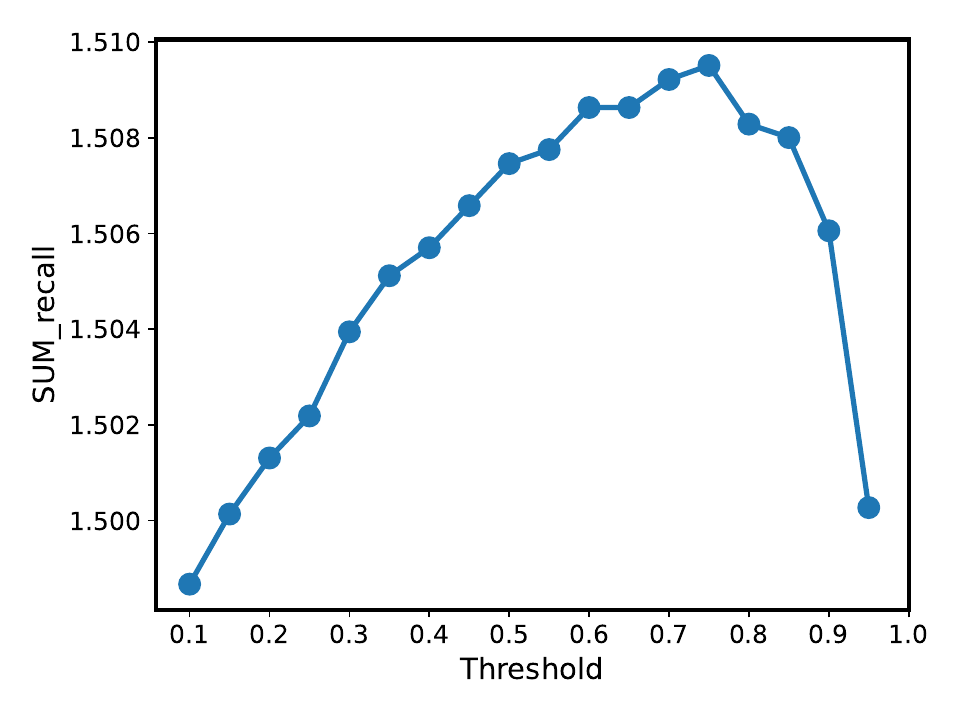}
        \label{fig:base_scena_test}
    }
    \caption{ODCT threshold selection on the validation and test sets.}
    \label{fig:threshold_optimize}
    \vspace{-2mm}
\end{figure}

\begin{figure*}[t]
    \centering
    \setlength{\tabcolsep}{3pt}
    \begin{tabular}{c c c}
        \includegraphics[width=0.32\textwidth]{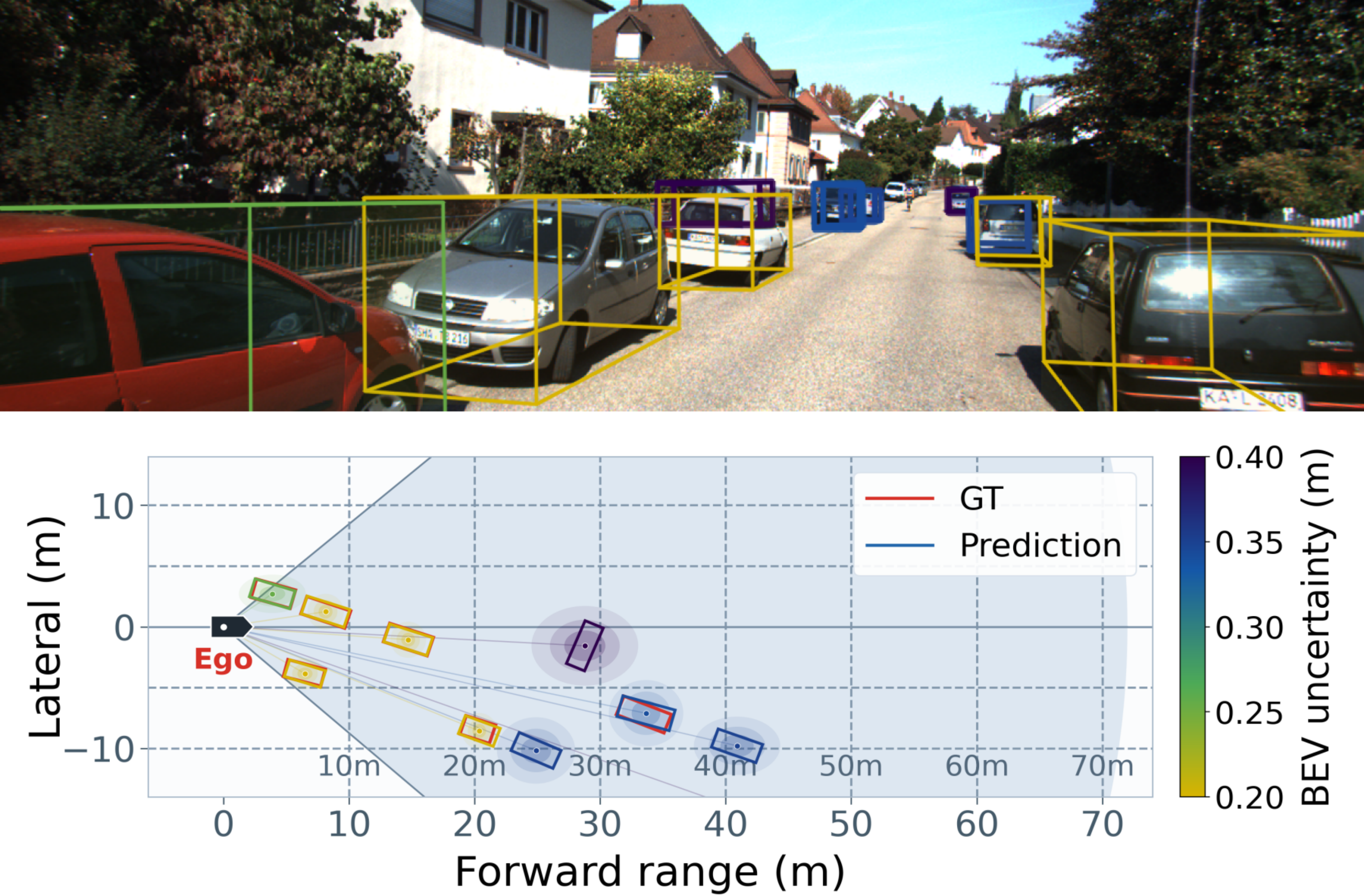} &
        \includegraphics[width=0.32\textwidth]{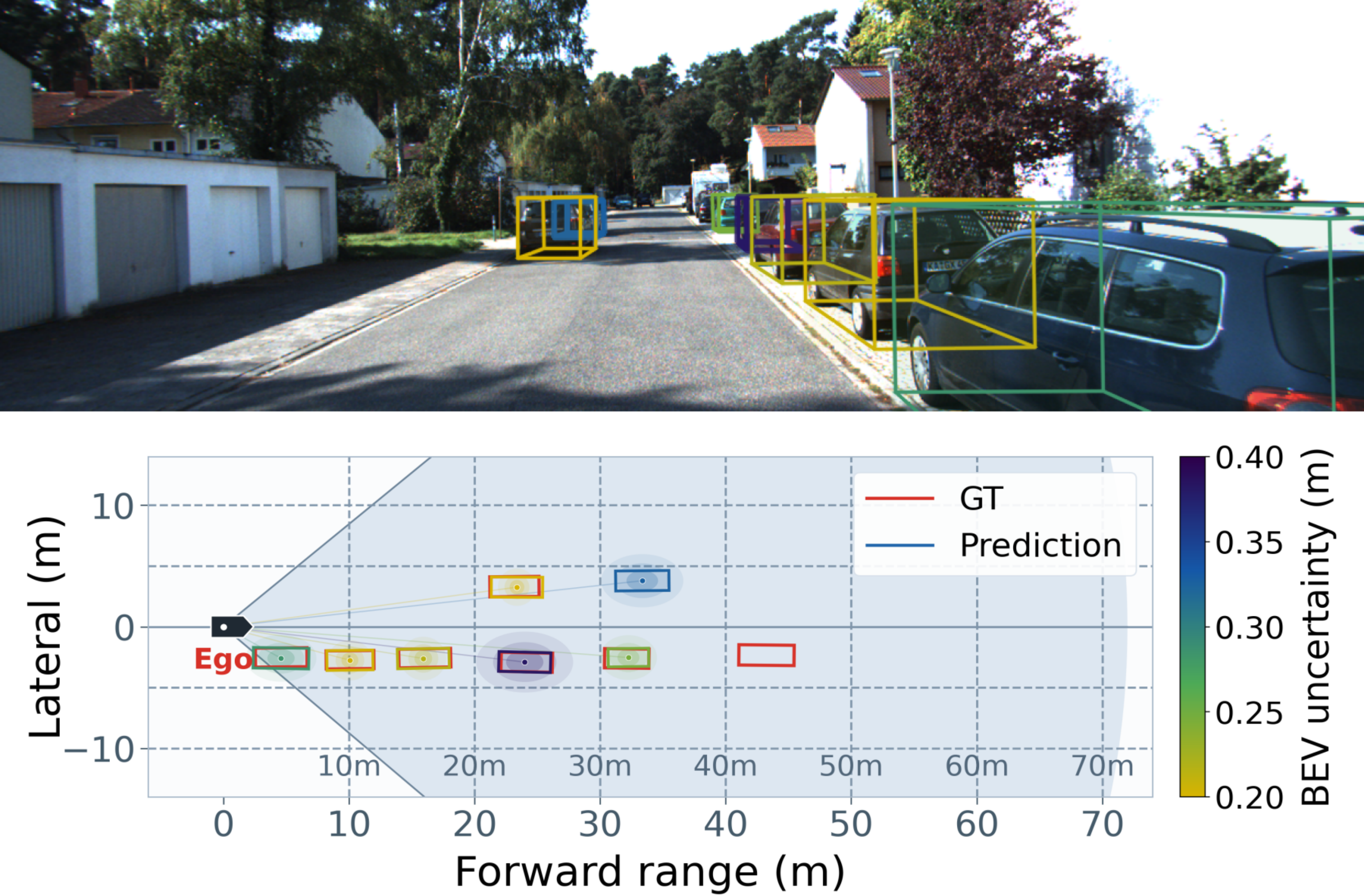} &
        \includegraphics[width=0.32\textwidth]{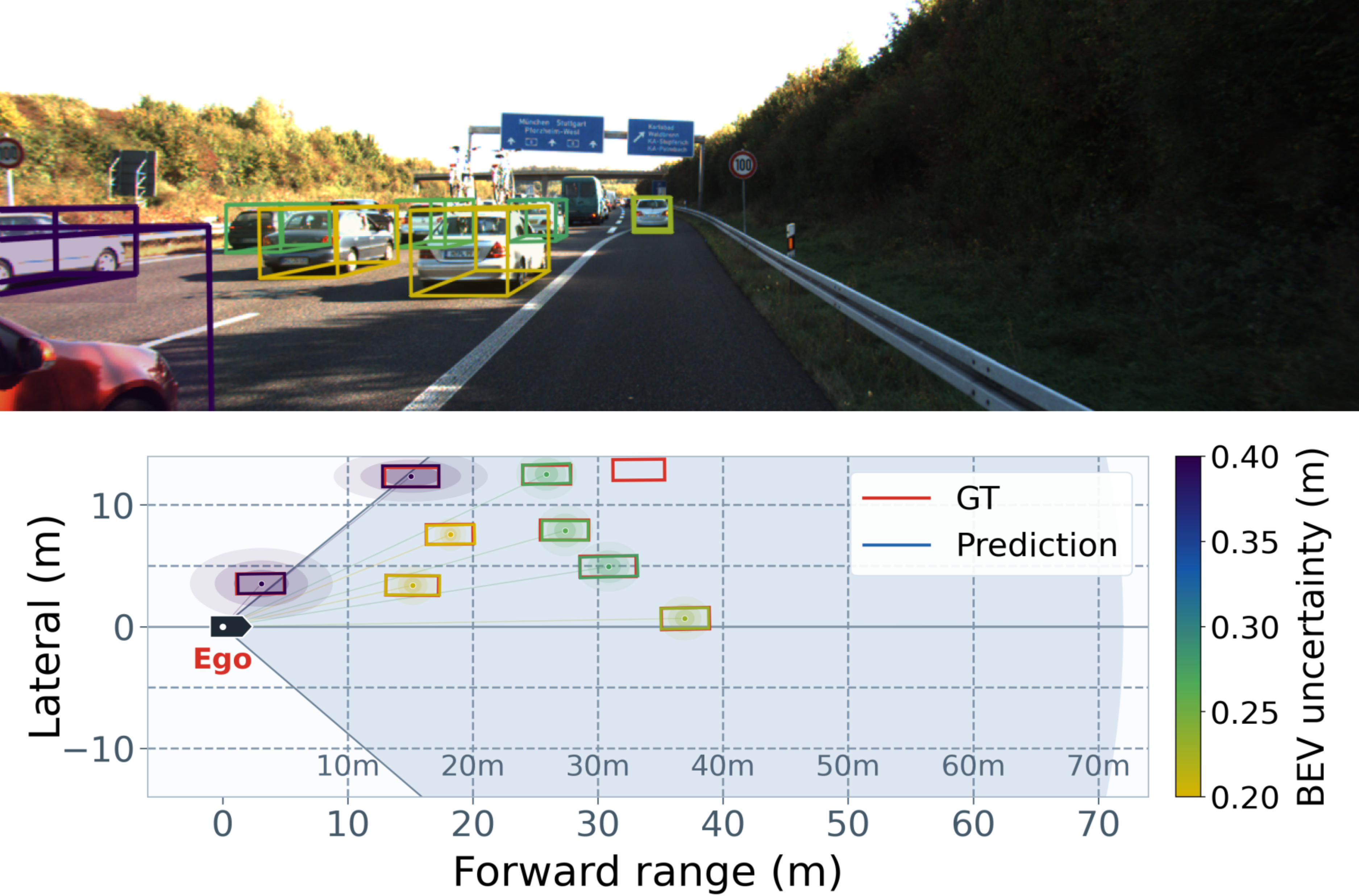} \\
        
        \includegraphics[width=0.32\textwidth]{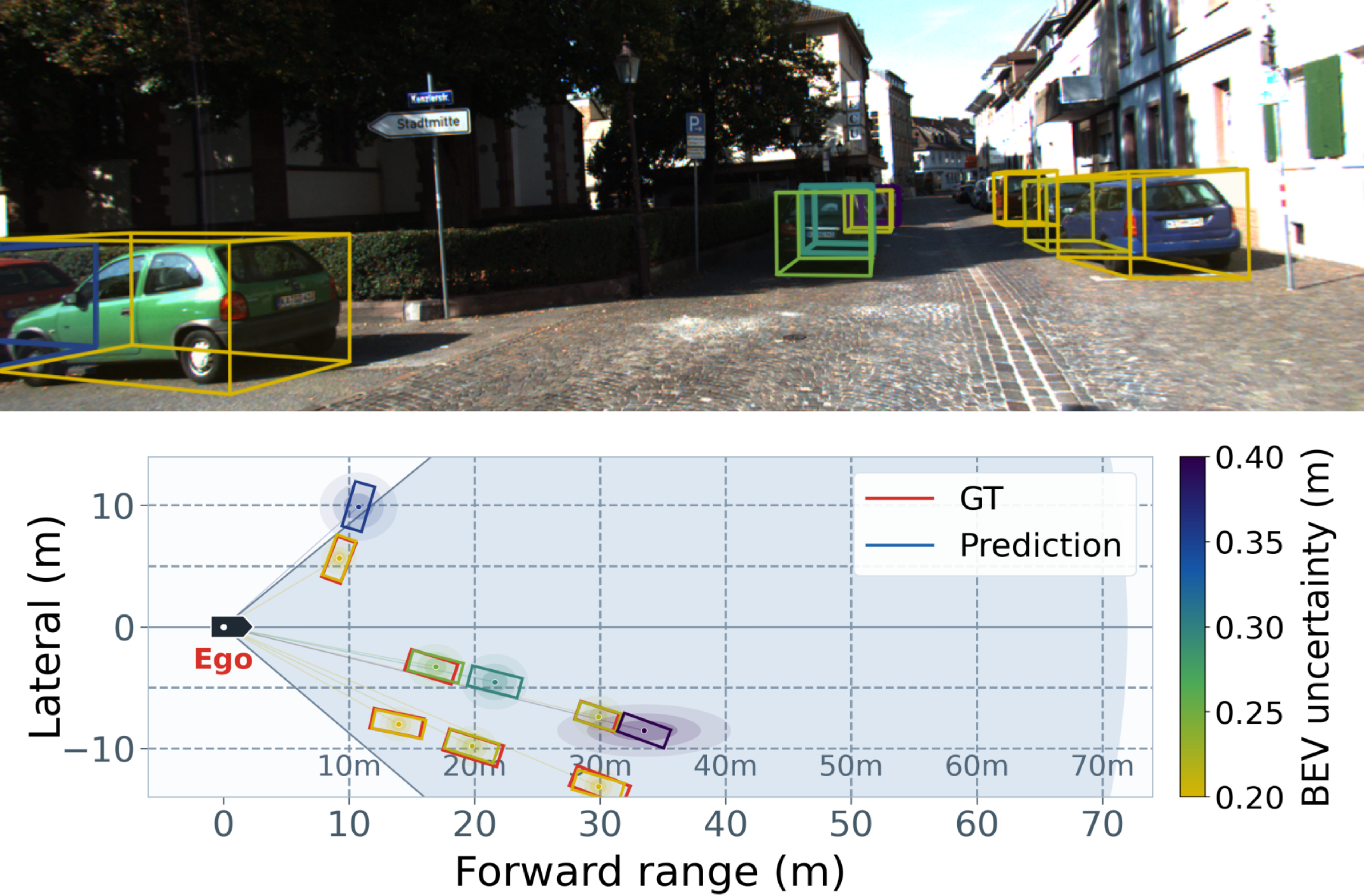} &
        \includegraphics[width=0.32\textwidth]{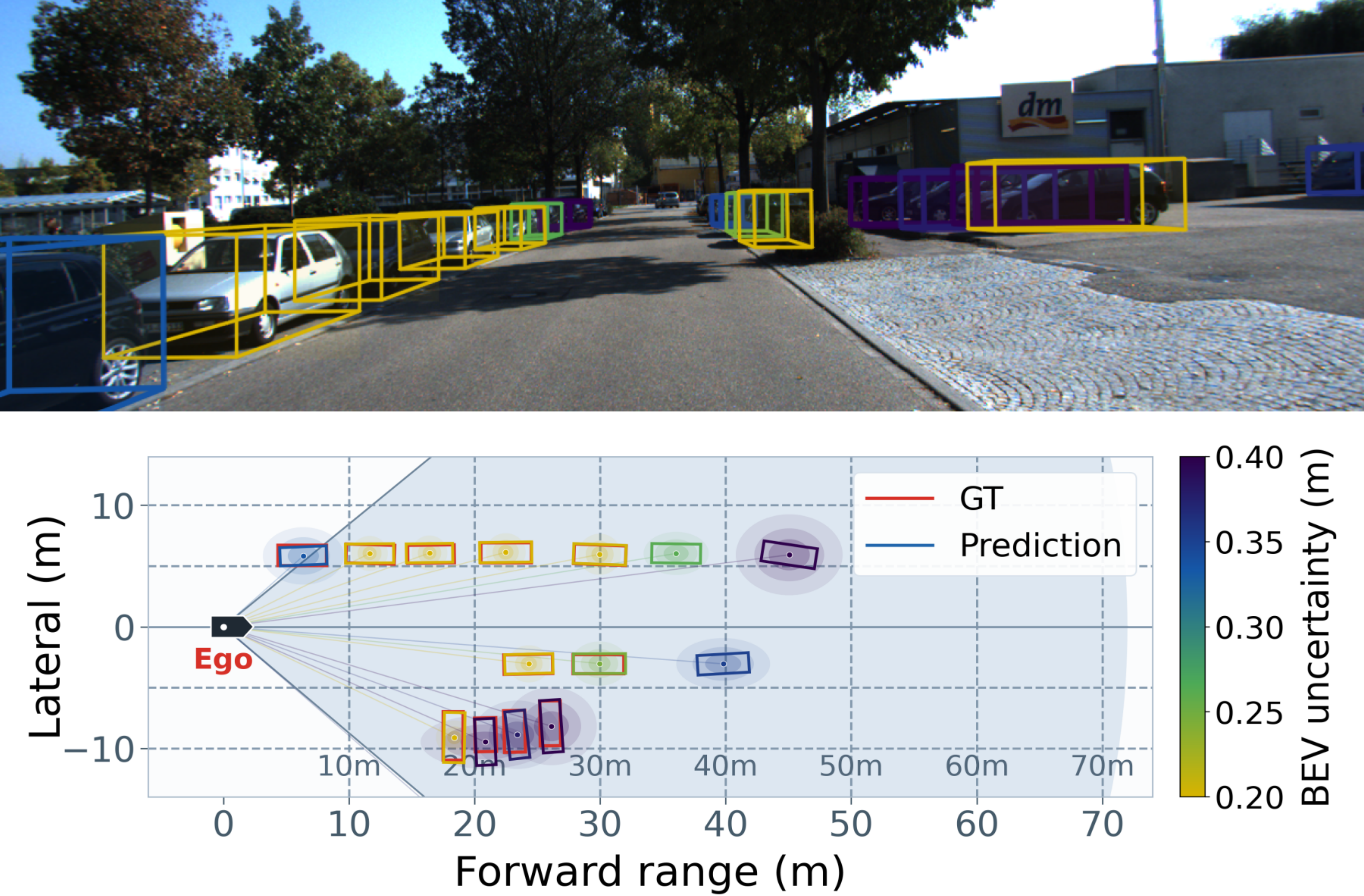} &
        \includegraphics[width=0.32\textwidth]{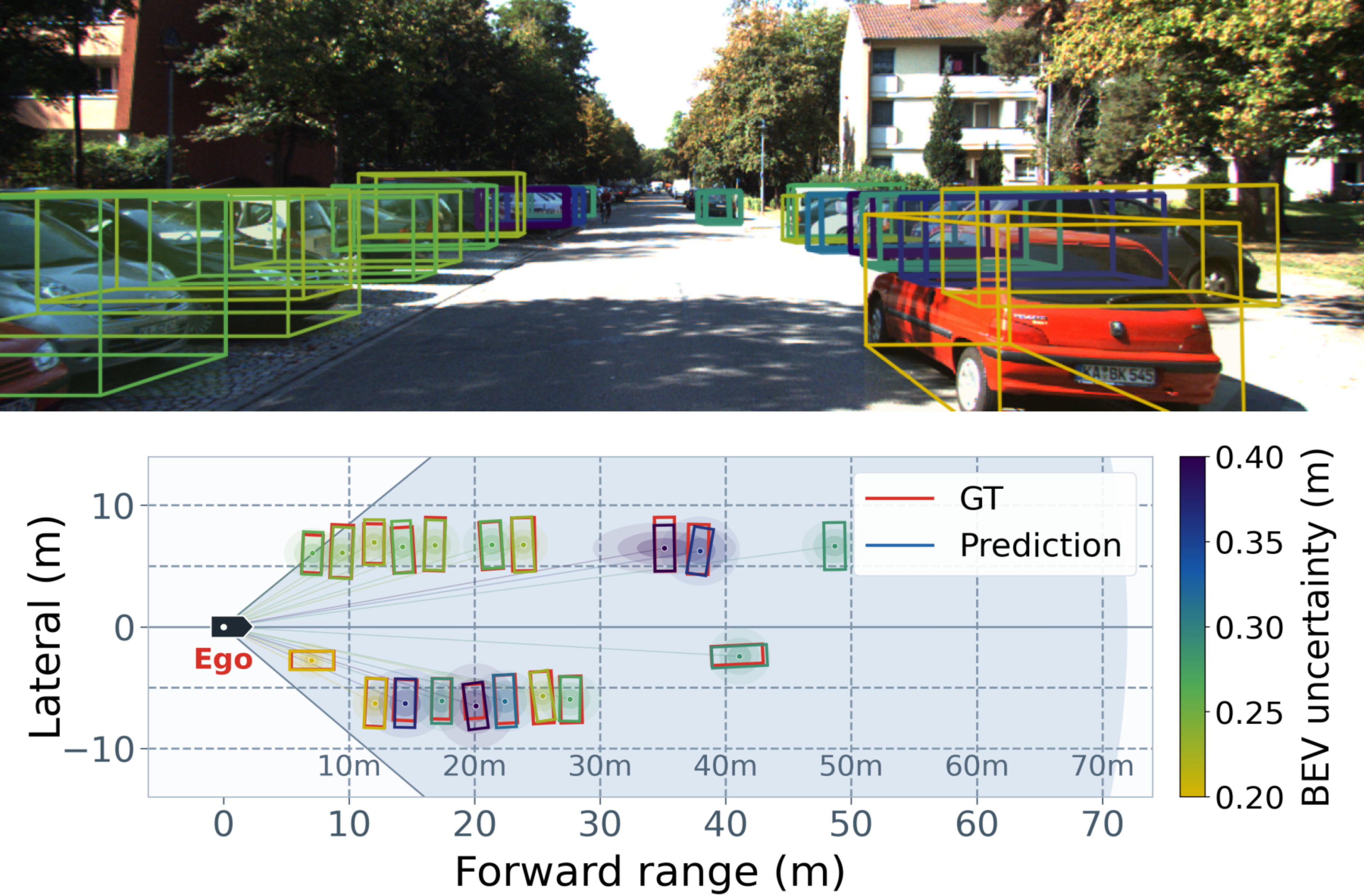}
    \end{tabular}
    \caption{Qualitative examples of 3D object detection with predicted geometric uncertainty.}
    \label{fig:kitti_demo_uncertainty}
    \vspace{-3.5mm}
\end{figure*}

\subsubsection{Semantic Uncertainty for Unknown-Object Awareness}

We conduct experiments to validate the LVM-based perception module and the ODCT-based semantic uncertainty screening mechanism. We collect a dataset in the CARLA Town06 map, generating over $30{,}000$ frames of $1080 \times 720$ RGB images synchronized with vehicle odometries. Each frame is annotated with ground-truth bounding boxes and undergoes quality filtering to ensure object visibility, with occluded or partially visible instances excluded during data curation. As shown in Table~\ref{tab:data_splits}, the dataset contains $49$ object categories (including vehicles and $48$ other object types in CARLA), and is partitioned into local and cloud subsets for open-vocabulary detection. Among these $49$ categories, $38$ are treated as known classes and the remaining $11$ as unknown classes. The local training set contains $9{,}647$ samples covering all $38$ known classes. The validation set contains $2{,}060$ samples with $21$ known classes and $5$ unknown classes, while the test set contains $3{,}400$ samples with $19$ known classes and the remaining $6$ unknown classes. The unknown classes in the validation and test subsets are non-overlapping and excluded from local-model training.

The local and cloud models are implemented based on the UnSniffer architecture~\cite{liang2023unknown}, which provides a confidence score for each detection. The local detection model is trained on the training subset and fine-tuned on the validation set to optimize the threshold $c_{\mathrm{th}}$. The evaluation is performed on the test set. For the cloud detector, a larger training set of $15{,}107$ samples spanning all $49$ predefined categories is used.

Fig.~\ref{fig:threshold_optimize} depicts the value of $G$ (i.e., the sum of the recall of in-distribution detections and the recall of out-of-distribution detections in \eqref{eq:recall}) versus $c_{\mathrm{th}}$. On the validation set, the recall score reaches its maximum at $c_{\mathrm{th}}=0.8$, while on the test set the highest score is achieved at $c_{\mathrm{th}}=0.75$. The small shift of the optimal threshold between the two splits suggests that confidence thresholding provides a stable criterion for semantic uncertainty screening, even though the validation and test sets contain non-overlapping unknown classes. Meanwhile, this result also reflects an inherent tradeoff in confidence-based unknown-object detection: a larger threshold tends to preserve more potentially unknown objects for cloud-side reasoning, but may also introduce more false alarms among known objects; a smaller threshold suppresses unnecessary warnings but risks missing ambiguous objects. The ODCT procedure therefore does not aim to make the local model perfectly recognize all unknown objects, but to choose a practical operating point for early screening.

With the optimized threshold, the local model achieves recalls of $0.83$ for known objects and $0.71$ for unknown objects on the validation set. On the test set, the known-object recall increases to $0.89$, while the unknown-object recall decreases to $0.61$. This decrease is expected because the test unknown classes are not seen during either local training or threshold selection, and may present different visual appearances from the validation unknown classes. Nevertheless, the local model still retains a considerable portion of unknown objects as low-confidence detections. This is sufficient for our purpose, since semantic uncertainty is used to identify candidate ambiguous cases rather than to provide the final semantic interpretation at the edge. The cloud model, trained on all predefined classes, achieves a recall of $0.89$ on the known-class validation set, confirming its stronger recognition capability under richer training coverage and cloud-side computation.

We further visualize representative perception results in Fig.~\ref{fig:semantic_compare}. Case~1 contains only known objects, while Cases~2--3 include unknown objects. In Case~1, the local model provides reliable detections, indicating that cloud reasoning would provide limited additional semantic benefit. In contrast, in Cases~2--3, the local model generates bounding boxes for both known and unknown objects, but fails to assign reliable semantic labels to unknown objects such as the trash can and vending machine. This behavior is useful for fast--slow collaboration: the local model does not need to fully understand every unknown object, but it should avoid silently ignoring potentially relevant objects. The low-confidence detections below $c_{\mathrm{th}}=0.8$ allow the onboard system to raise an ``unknown-object'' warning and mark these objects as candidate cases for cloud-assisted reasoning. The cloud model then recognizes the unknown objects and further segments the image into lane and object masks, allowing the system to recover object--lane topology for downstream decision making. These results validate semantic uncertainty as an effective screening signal: it separates normal local-perception cases from ambiguous cases that may deserve cloud reasoning, while leaving the final decision of whether cloud assistance is actually useful to the downstream planning-gain assessment.

\begin{figure}[t]
    \vspace{-1mm}
    \centering
    \subfloat[Bin statistics]{
        \includegraphics[width=0.44\columnwidth]{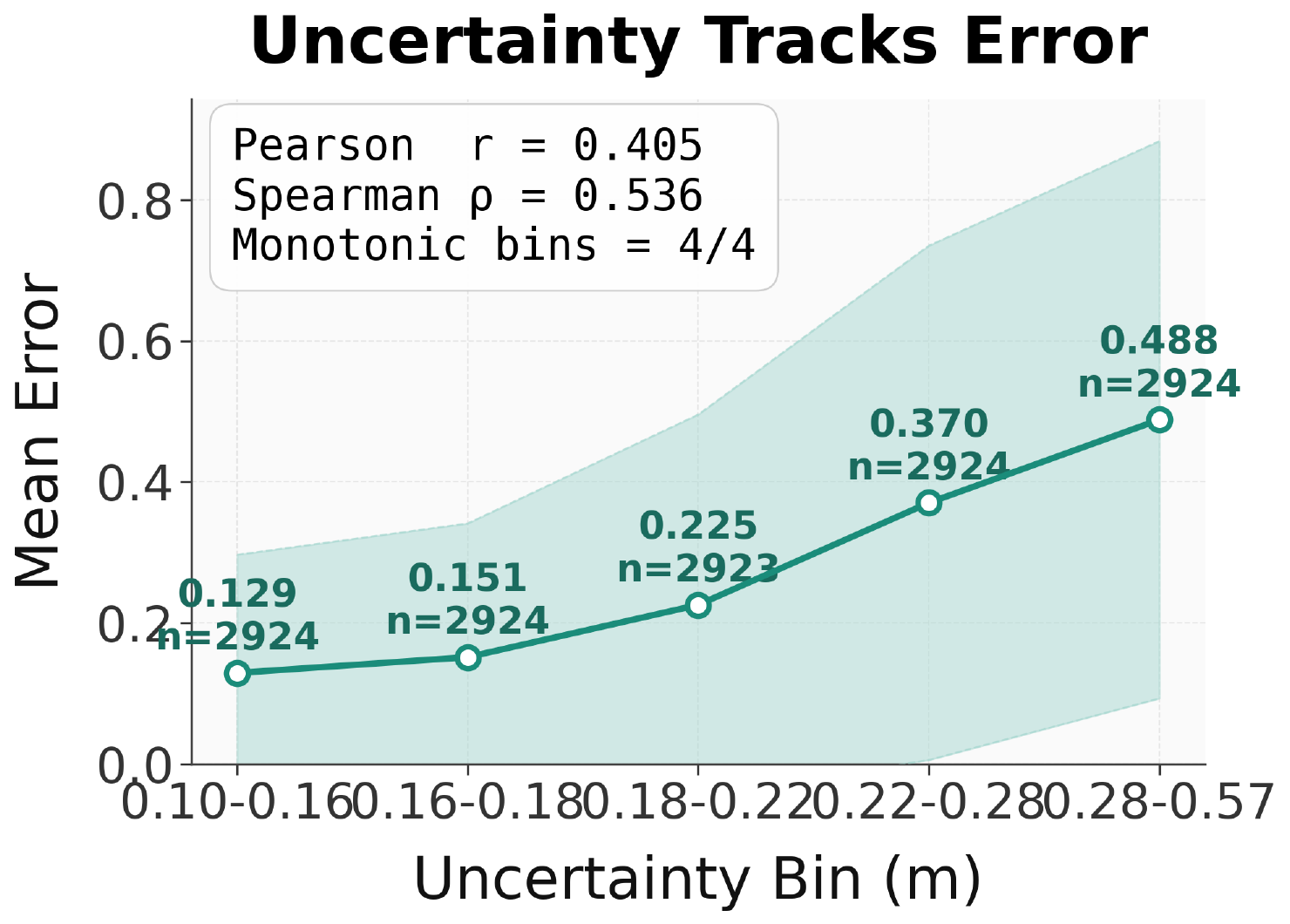}
        \label{fig:uncertainty_error_bin}
    }
    \hfil
    \subfloat[Scatter distribution]{
        \includegraphics[width=0.44\columnwidth]{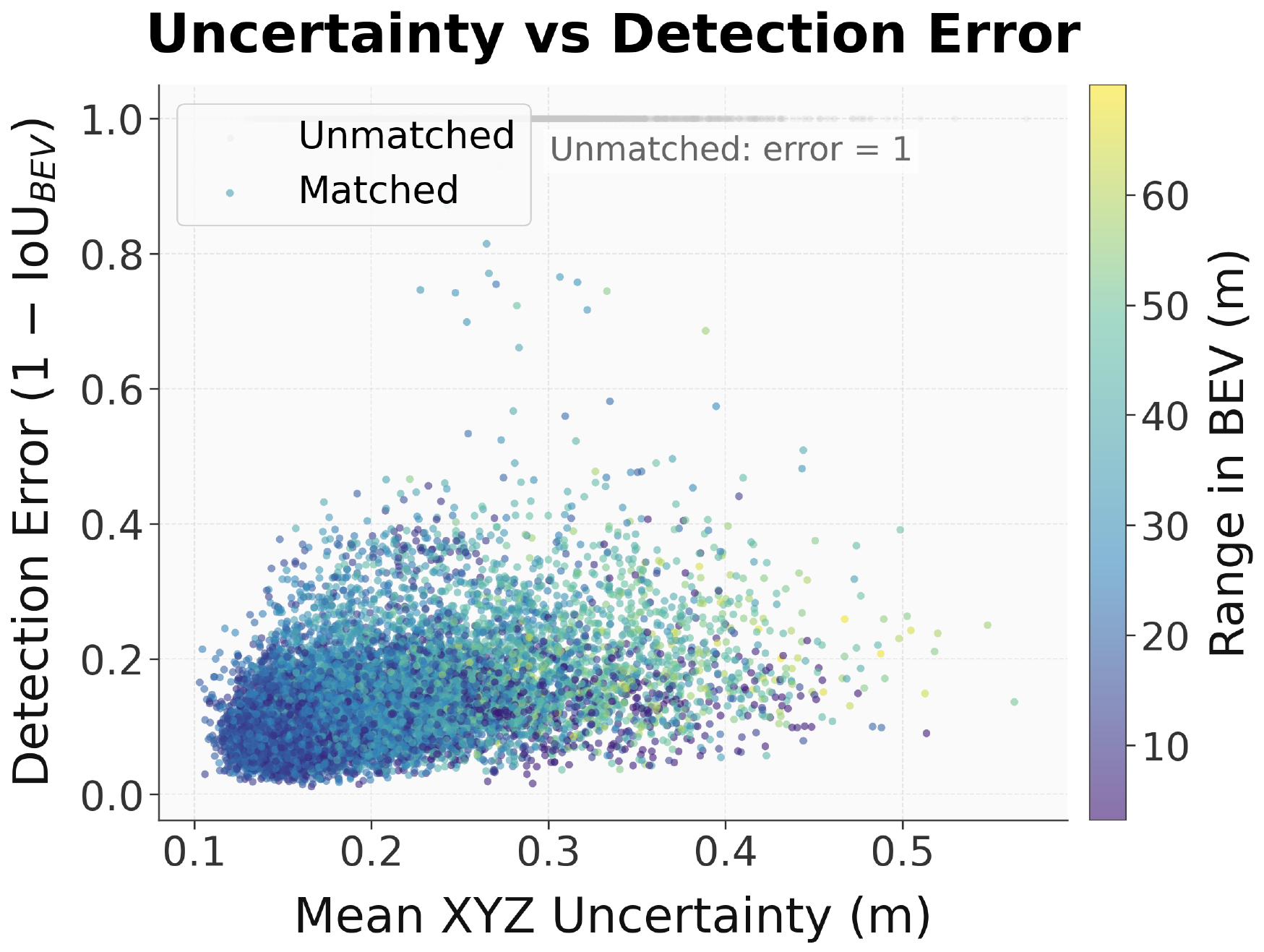}
        \label{fig:uncertainty_error_scatter}
    }
    \caption{Predicted geometric uncertainty versus localization error on KITTI.}
    \label{fig:uncertainty_error}
    \vspace{-2mm}
\end{figure}

\subsubsection{Geometric Uncertainty for 3D Object Localization}

We validate whether the proposed geometric uncertainty modeling reflects true localization ambiguity in 3D perception on the KITTI validation set, using the Gaussian uncertainty-aware 3D detector introduced in Section~\ref{subsec:probabilistic_obstacle}. The detector predicts both 3D bounding box parameters and their associated uncertainty, which are later used for planning-oriented obstacle inflation. For analysis, we use the mean predicted standard deviation over the spatial dimensions $(x,y,z)$ as a scalar uncertainty measure, and quantify localization error by $1-\mathrm{IoU}_{\mathrm{BEV}}$.

Fig.~\ref{fig:uncertainty_error_bin} presents the main quantitative result, where predictions are grouped into uncertainty bins and the average localization error is computed for each bin. The results show a clear monotonic trend: as predicted uncertainty increases, the average localization error rises from $0.129$ to $0.488$. This indicates that the uncertainty head tends to assign larger uncertainty to objects that are indeed harder to localize. The positive Pearson correlation of $0.405$ and Spearman correlation of $0.536$ further support this trend. Meanwhile, the scatter distribution in Fig.~\ref{fig:uncertainty_error_scatter} shows that the relationship is not perfectly deterministic at the instance level, which is expected because 3D localization error can also be affected by factors such as object distance, viewpoint, occlusion, object size, and LiDAR sparsity. Therefore, the predicted uncertainty should be interpreted as a statistical indicator of localization difficulty rather than an exact error predictor for every individual sample.

\begin{figure}[!t]
    \centering
    \subfloat[Distance vs. uncertainty]{
        \includegraphics[width=0.43\columnwidth]{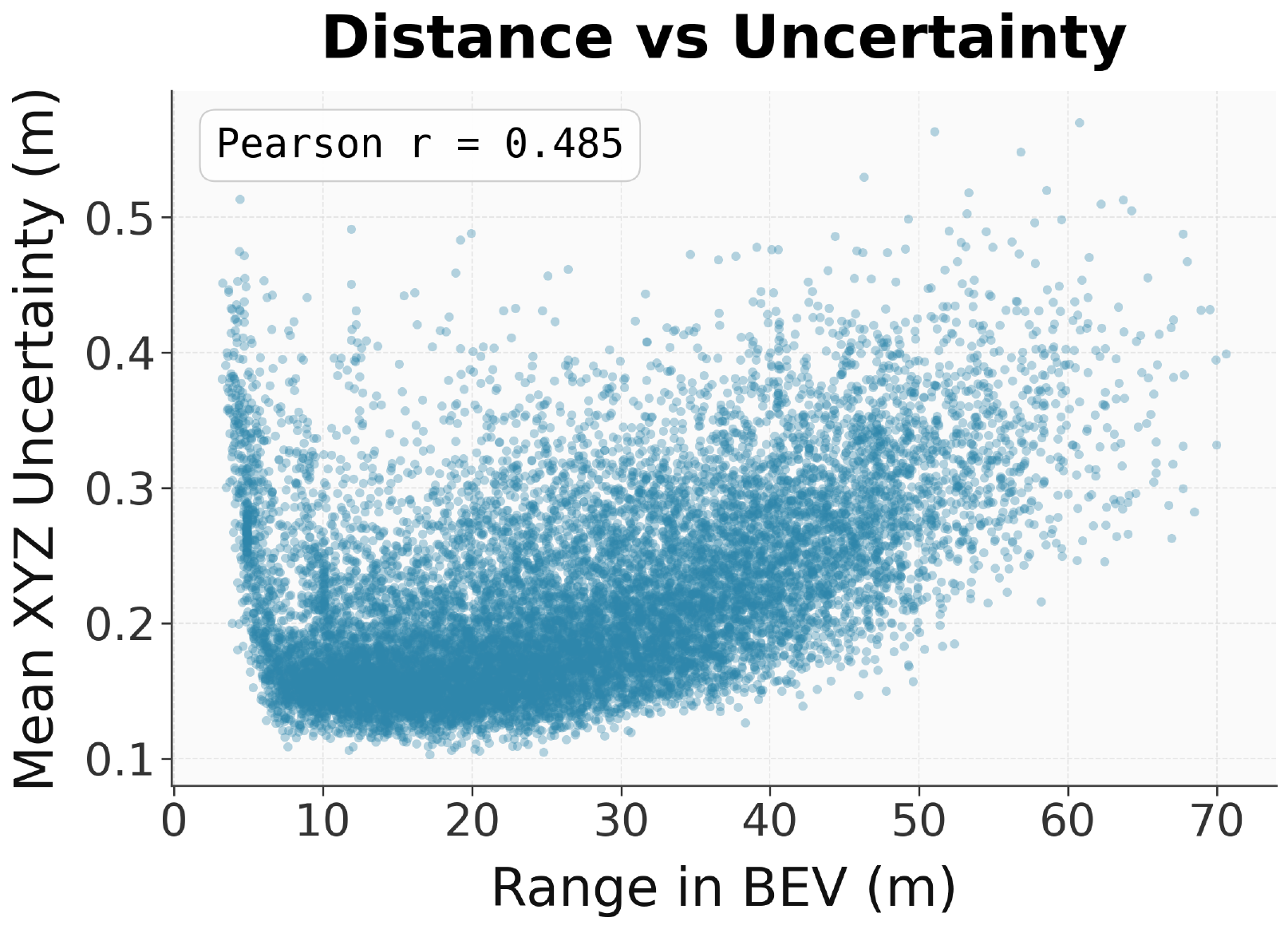}
        \label{fig:distance_uncertainty}
    }
    \hfil
    \subfloat[Distance vs. error]{
        \includegraphics[width=0.43\columnwidth]{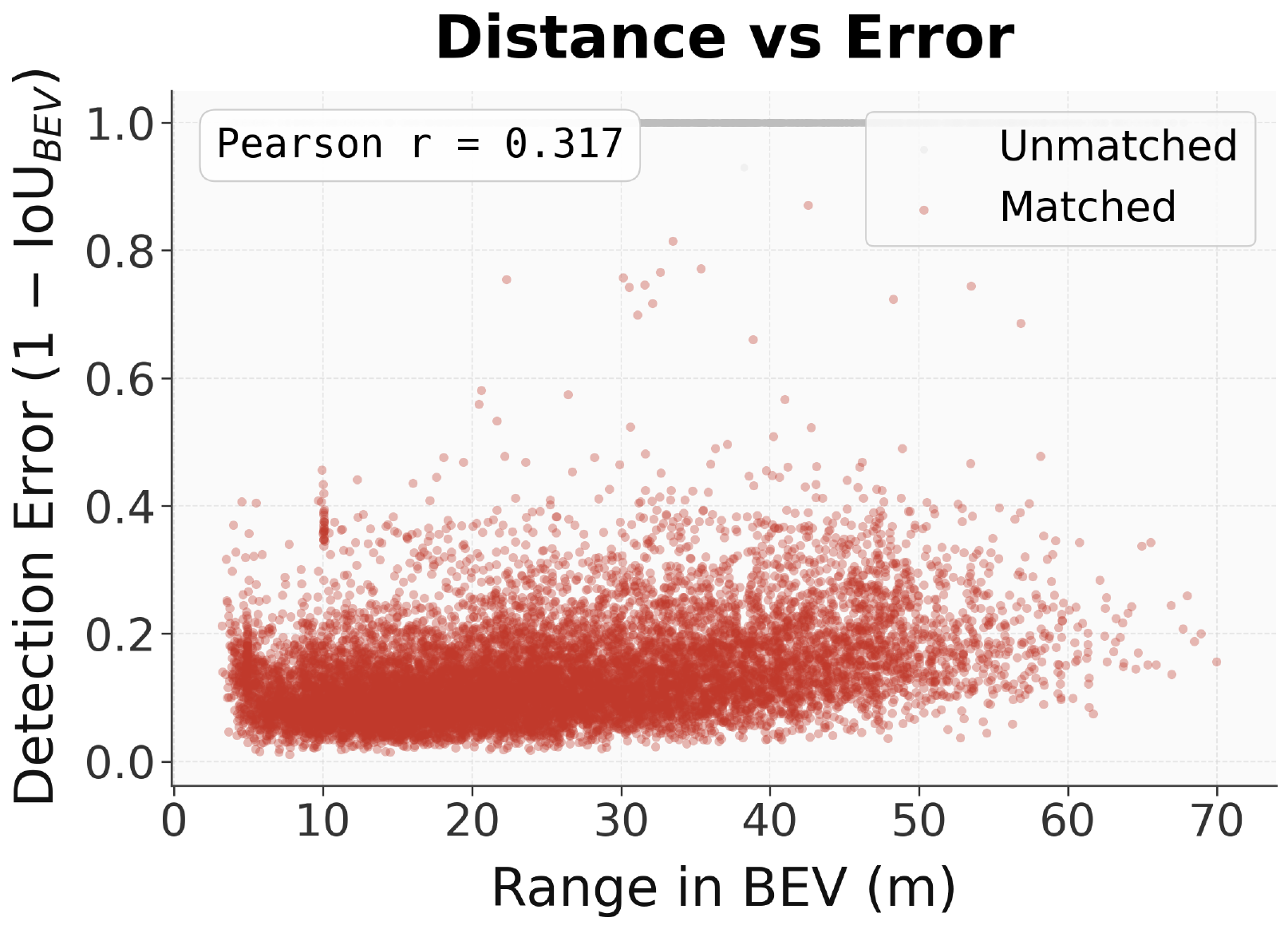}
        \label{fig:distance_error}
    }
    \caption{Relationship with predicted uncertainty, and localization error.}
    \label{fig:distance_uncertainty_error}
    \vspace{-2mm}
\end{figure}

Representative KITTI examples in Fig.~\ref{fig:kitti_demo_uncertainty} provide qualitative evidence consistent with this interpretation. Both image-view and BEV-view projections show larger predicted uncertainty in visually challenging cases, such as distant objects, sparse observations, or ambiguous object extents. Fig.~\ref{fig:distance_uncertainty_error} further explains this phenomenon from a physical perspective. As object range increases, image details become less discriminative and LiDAR points become sparser, making accurate 3D localization more difficult. Consequently, both predicted uncertainty and actual localization error tend to increase with distance, which matches the intuitive behavior of 3D perception.

For completeness, Table~\ref{tab:kitti_gaussian_det} reports the standard KITTI detection accuracy of the uncertainty-aware detector. The results show that the detector maintains competitive 3D detection performance while producing meaningful uncertainty estimates. This is important because the uncertainty branch should provide additional planning-relevant information without substantially compromising the original detection capability. Overall, these results show that the learned geometric uncertainty is not arbitrary: it follows the expected difficulty pattern of 3D localization and captures meaningful spatial ambiguity. Therefore, it provides a reasonable uncertainty representation for obstacle inflation and subsequent simulation-in-the-loop planning-gain assessment.

\begin{table}[!t]
\centering
\caption{KITTI detection accuracy of the uncertainty-aware detector.}
\label{tab:kitti_gaussian_det}
\setlength{\tabcolsep}{5pt}
\renewcommand{\arraystretch}{1.2}
\begin{tabular}{lccc}
\toprule
\textbf{Metric} & \textbf{Easy} & \textbf{Moderate} & \textbf{Hard} \\
\midrule
3D AP@0.70      & 89.48 & 79.20 & 78.66 \\
BEV AP@0.70     & 90.33 & 88.15 & 87.71 \\
BBox AP@0.70    & 98.61 & 89.60 & 89.24 \\
\midrule
3D AP\_R40@0.70 & 92.83 & 83.57 & 81.14 \\
BEV AP\_R40@0.70& 96.09 & 89.65 & 87.24 \\
BBox AP\_R40@0.70 & 99.27 & 95.33 & 92.95 \\
\bottomrule
\end{tabular}
\vspace{-2mm}
\end{table}

\subsection{Experiment 2: Scenarios with Static Obstacles}

\subsubsection{System-Level Performance Comparison}

\begin{figure}[!t]
    \centering
    \includegraphics[width=0.9\columnwidth]{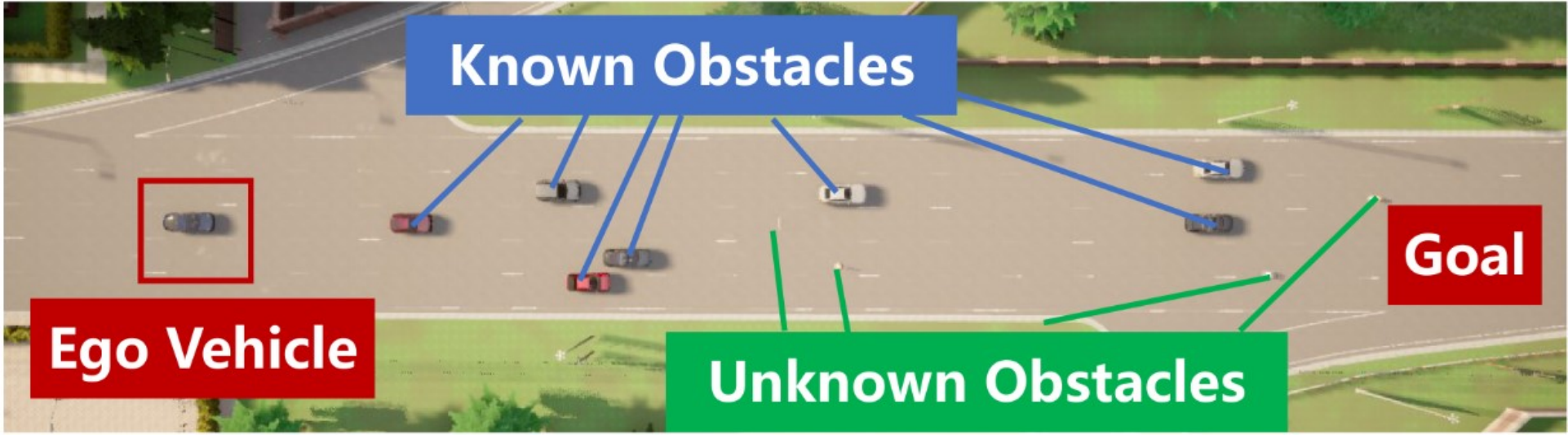}
    \caption{Static-obstacle scenario configuration for Experiment~2.}
    \label{figure:setting}
    \vspace{-2mm}
\end{figure}

Next, we evaluate the LOS, PCS, and SIGMA schemes at a target speed of 6\,m/s in the CARLA Town06 map.  As shown in Fig.~\ref{figure:setting}, the distance between the starting and goal positions is \rev{approximately} $110$ meters.  A total of $7$ known objects (vehicles) and $4$ unknown objects (two traffic cones, one street barrier, and one trash can) are placed along the route.  These unknown objects form two clusters, creating two navigation critical regions.

\begin{figure*}[t]
    \centering
    \subfloat[LOS.]{
        \includegraphics[width=0.31\textwidth]{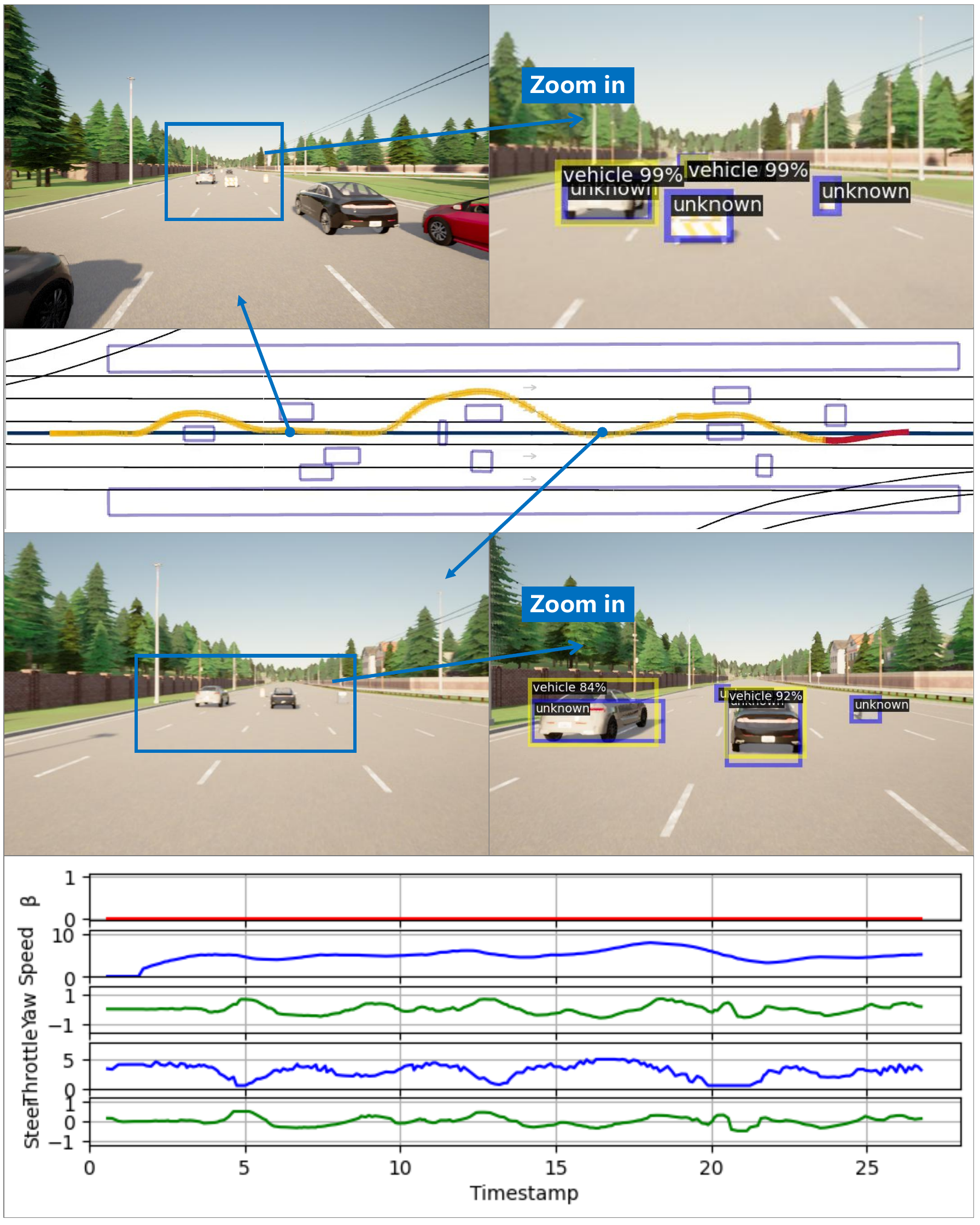}
        \label{fig:exp2_los}
    }
    \hfil
    \subfloat[PCS.]{
        \includegraphics[width=0.31\textwidth]{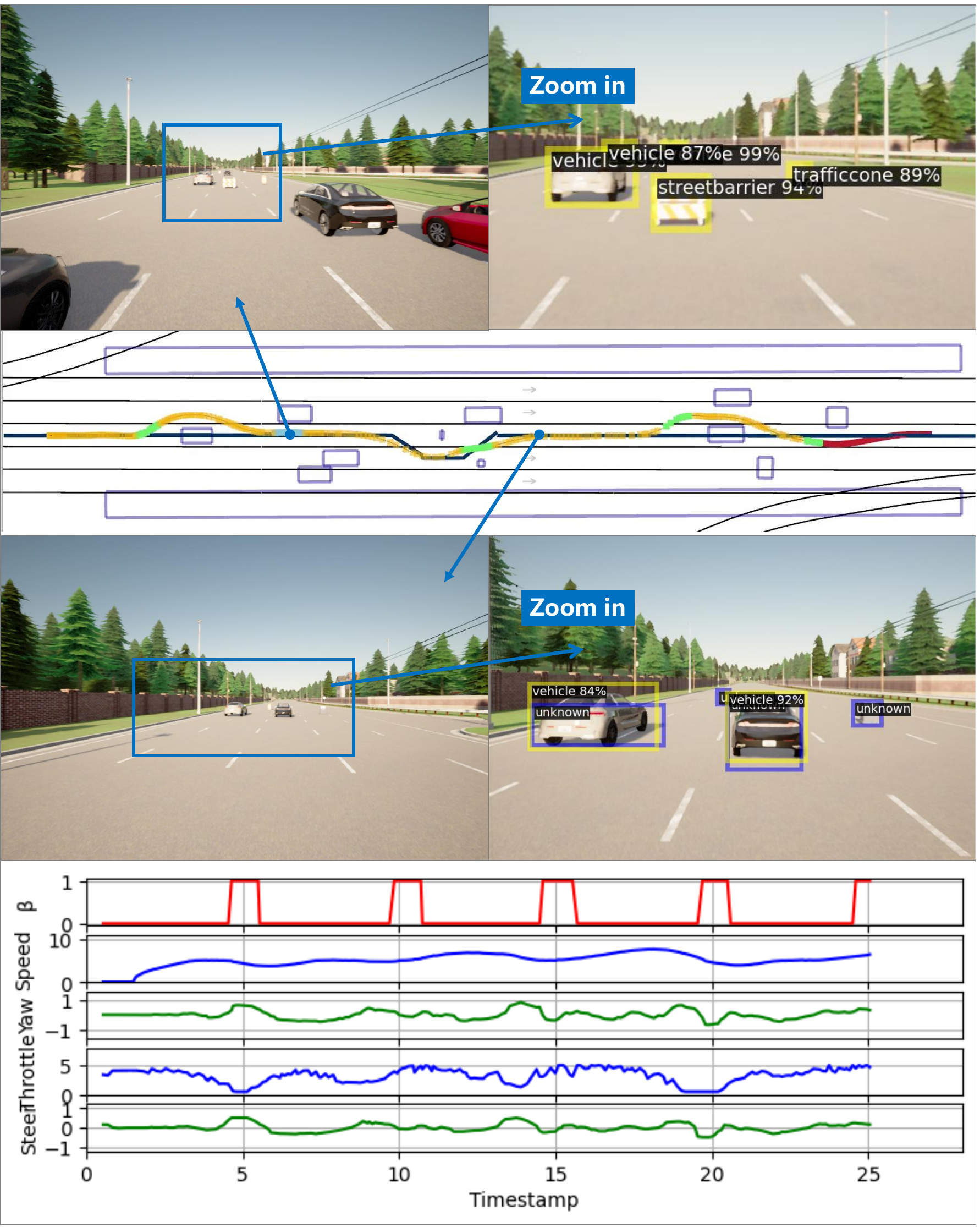}
        \label{fig:exp2_pcs}
    }
    \hfil
    \subfloat[SIGMA (Ours).]{
        \includegraphics[width=0.31\textwidth]{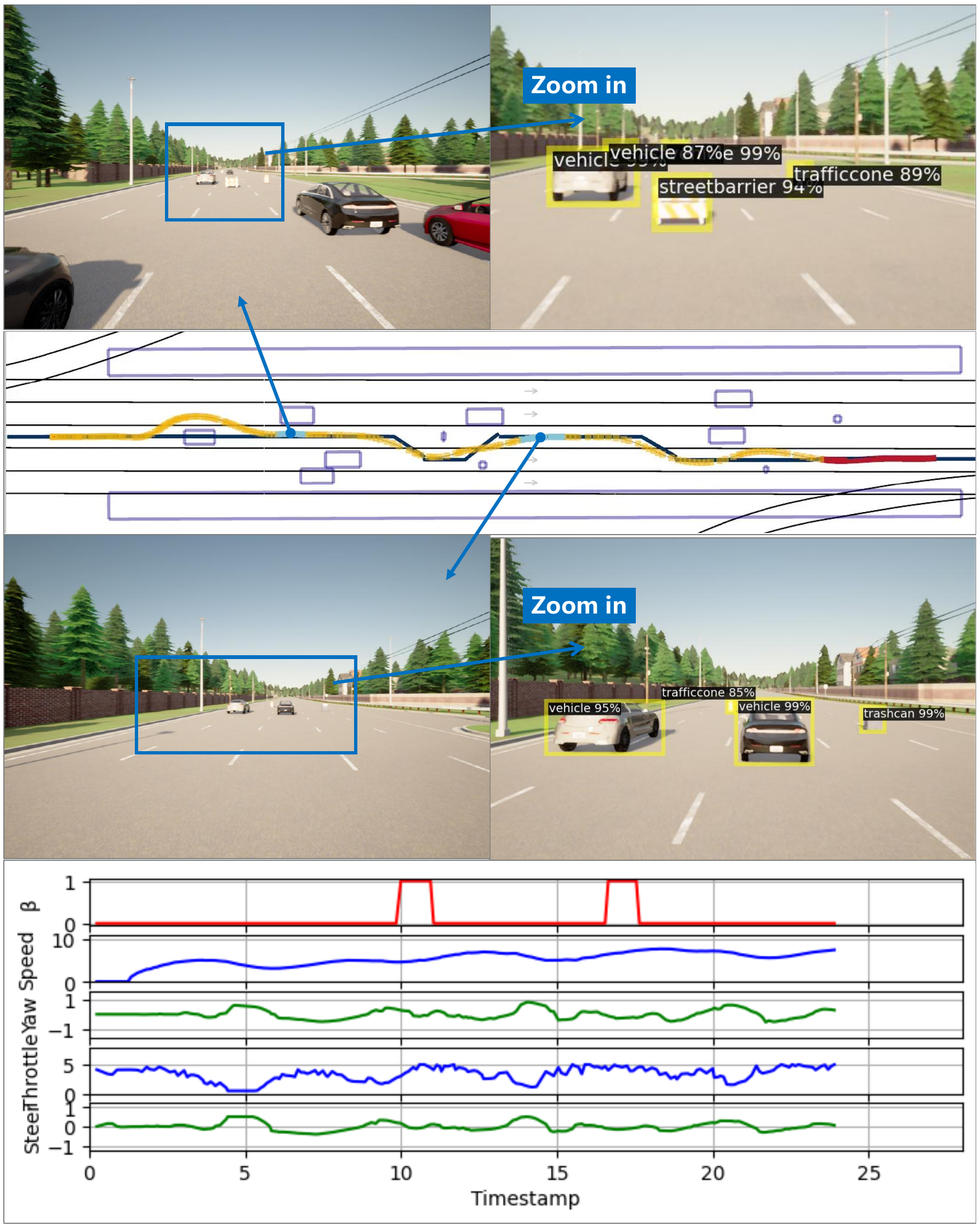}
        \label{fig:exp2_sigma}
    }
    \caption{Static-scenario experimental results: trajectories, control profiles, and collaboration states.}
    \label{fig:experiment1}
    \vspace{-1mm}
\end{figure*}

The trajectories, control profiles, and collaboration states (i.e., $\{\beta_t\}$) of different schemes are shown in Fig.~\ref{fig:experiment1}.  Yellow trajectories represent local navigation, while blue trajectories indicate cloud-guided navigation.  Green segments denote cloud engagement without trajectory variations, and red curves represent predicted MPC rollouts. Without cloud collaboration, the LOS scheme conservatively treats the unknown objects as obstacles and avoids them, resulting in the longest trajectory and finish time.  The PCS scheme, which relies on a fixed cloud-interaction intervals (5\,s), handles the first obstacle cluster but fails to react optimally to the second one, leading to suboptimal detours.  In contrast, the SIGMA scheme dynamically triggers cloud interaction based on ODCT-based semantic screening (Section~\ref{subsec:2d_semantic_ood}) and the EPG-guided collaboration strategy, handling both obstacle clusters and achieving the shortest trajectory and finish time.

More importantly, SIGMA reduces the number of cloud interactions by $50\%$ compared with PCS (2 vs.\ 4 times).  This improvement is achieved by avoiding unnecessary cloud queries at non-critical locations (green segments in PCS), where improved perception would not lead to meaningful trajectory changes.  This demonstrates the \textbf{resource efficiency} of the proposed collaboration strategy. Additionally, at the second obstacle cluster, the ego vehicle initially deviates due to incomplete perception but quickly adjusts its trajectory once cloud guidance becomes available.  This behavior highlights the \textbf{self-correction capability} enabled by the integration of large vision models and MPC.

\begin{figure}[!t]
    \centering
    \includegraphics[width=0.9\columnwidth]{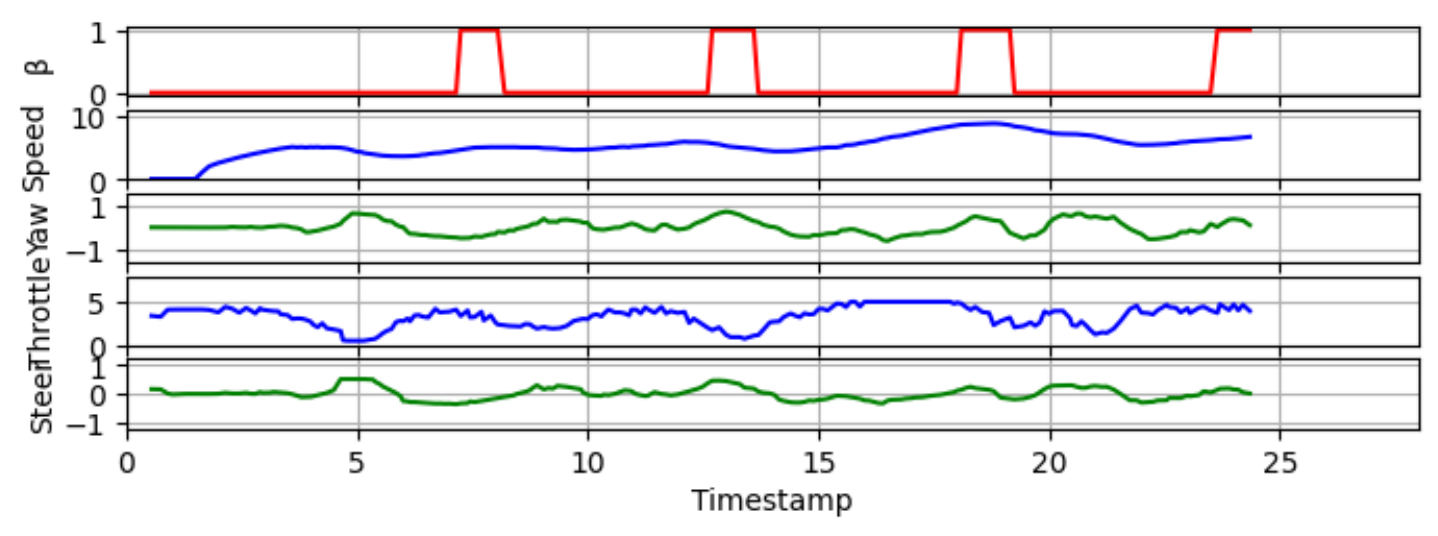}
    \vspace{-3mm}
    \caption{Trajectory and control profiles of PCS-2.}
    \label{figure:pcs2}
    \vspace{-1mm}
\end{figure}

\begin{table}[!t]
\centering
\caption{Quantitative comparison in Experiment~2.}
\label{table:params_2}
\setlength{\tabcolsep}{4pt}
\renewcommand{\arraystretch}{1.1}
\scalebox{1.0}{
\begin{tabular}{lccccc}
\toprule
\textbf{Method} & \textbf{FTime} & \textbf{TLen} & \textbf{AvgLD} & \textbf{SVar} & \textbf{MLat} \\
\textbf{Unit} & (s) & (m) & (m) & (m/s) & (m) \\
\midrule
LOS    & 25.94 & 124.54 & 0.53 & 1.09 & 1.52 \\
PCS    & 24.09 & 122.40 & 0.59 & 1.13 & 1.89 \\
PCS-2  & 23.53 & 123.84 & 0.57 & 1.39 & 1.69 \\
SIGMA  & 22.35 & 121.85 & 0.60 & 1.30 & 1.76 \\
\bottomrule
\end{tabular}
}
\end{table}

A quantitative comparison of the schemes is shown in Table~\ref{table:params_2}. We repeat the experiment 10 times and report the average metrics. This evaluation includes finish time (FTime) in seconds, trajectory length (TLen) in meters, average lateral deviation (AvgLD) in meters, speed variability (SVar) in m/s, and maximum lateral deviation (MLat) in meters. To ensure a fair comparison, we also simulate a variant of PCS, termed PCS-2, which adopts a fixed cloud-interaction interval of 5.5\,s. Its trajectories and control profiles are shown in Fig.~\ref{figure:pcs2}. The proposed SIGMA achieves the smallest FTime and TLen, benefiting from cloud-guided LVM--MPC collaboration and EPG-guided collaboration timing.

\subsubsection{Simulation-in-the-Loop Forward Simulation Analysis}

\begin{figure*}[t]
    \centering
    \setlength{\tabcolsep}{3pt}
    \begin{tabular}{c c c}
        \includegraphics[width=0.32\textwidth]{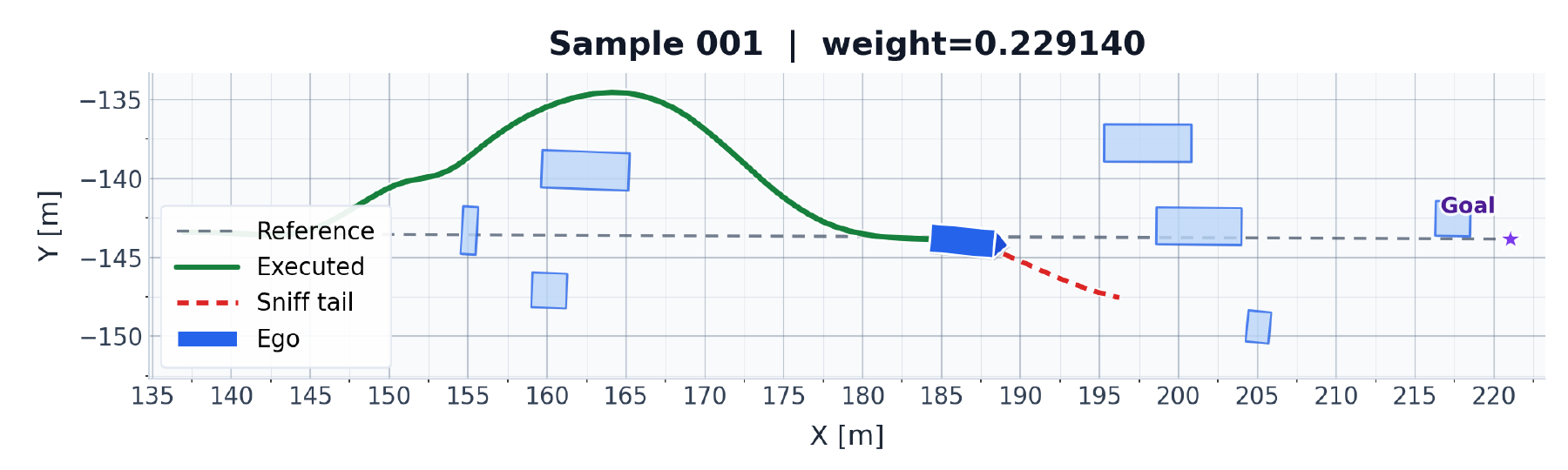} &
        \includegraphics[width=0.32\textwidth]{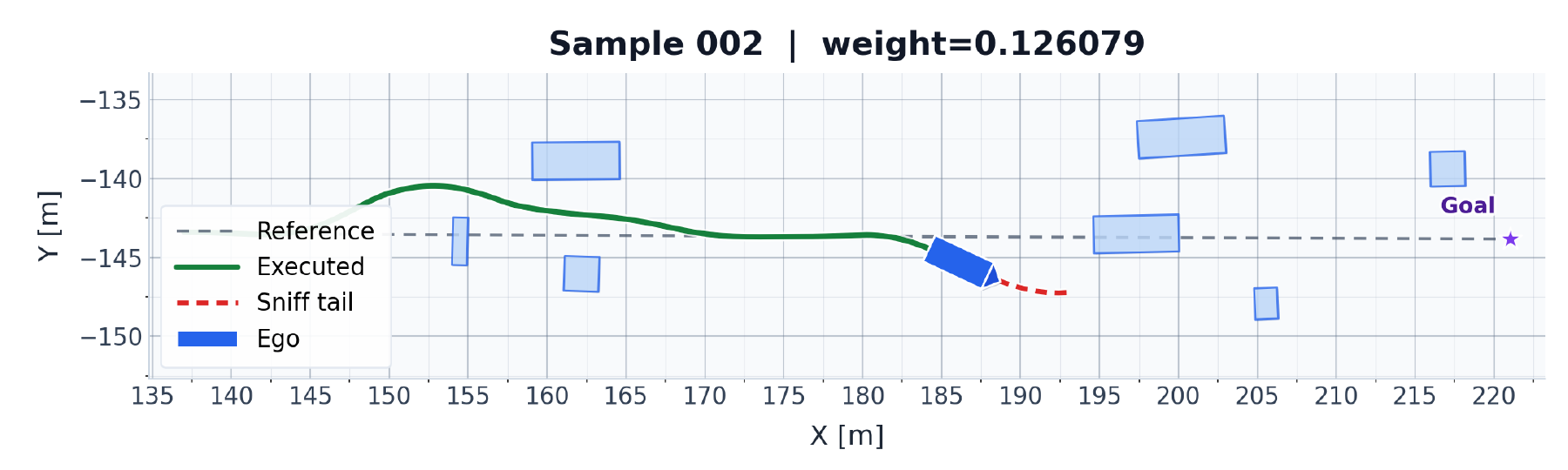} &
        \includegraphics[width=0.32\textwidth]{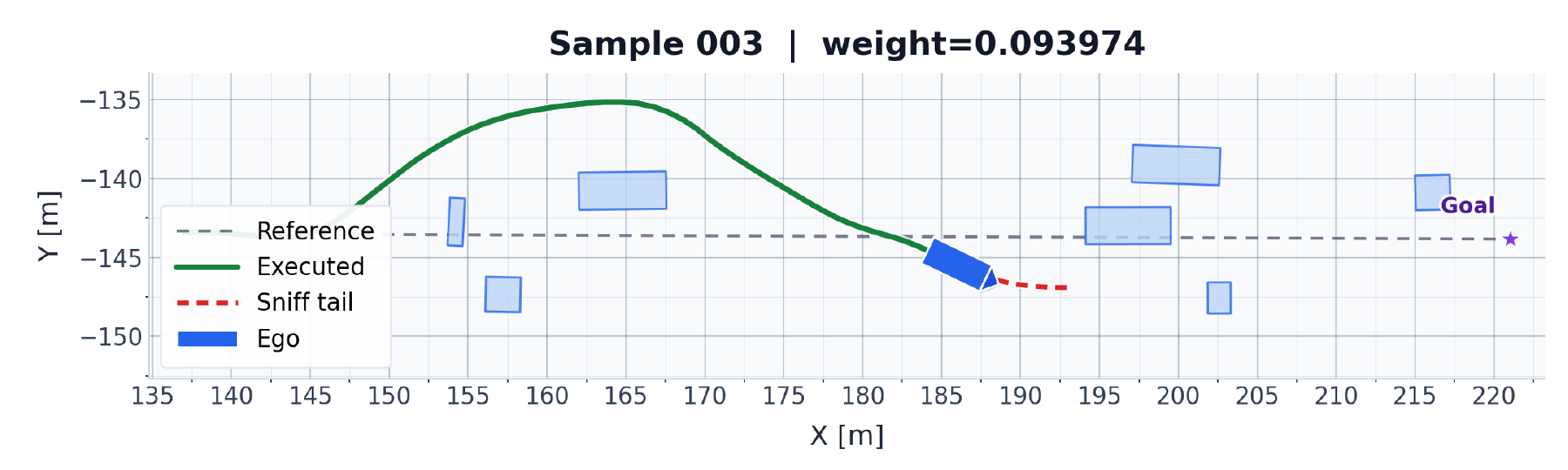} \\

        \includegraphics[width=0.32\textwidth]{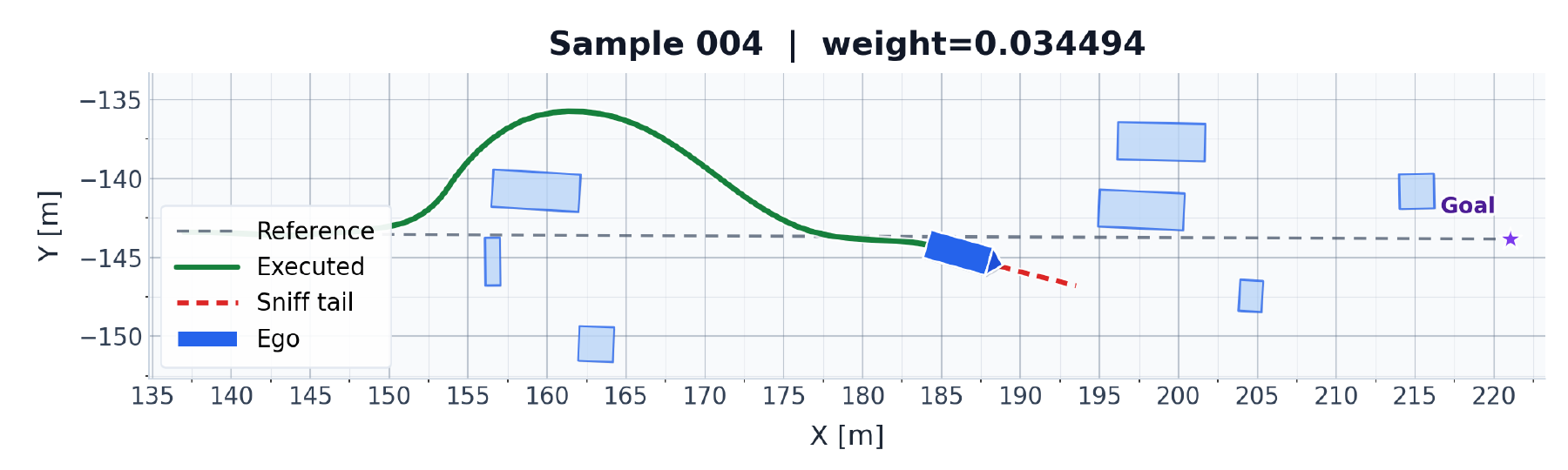} &
        \includegraphics[width=0.32\textwidth]{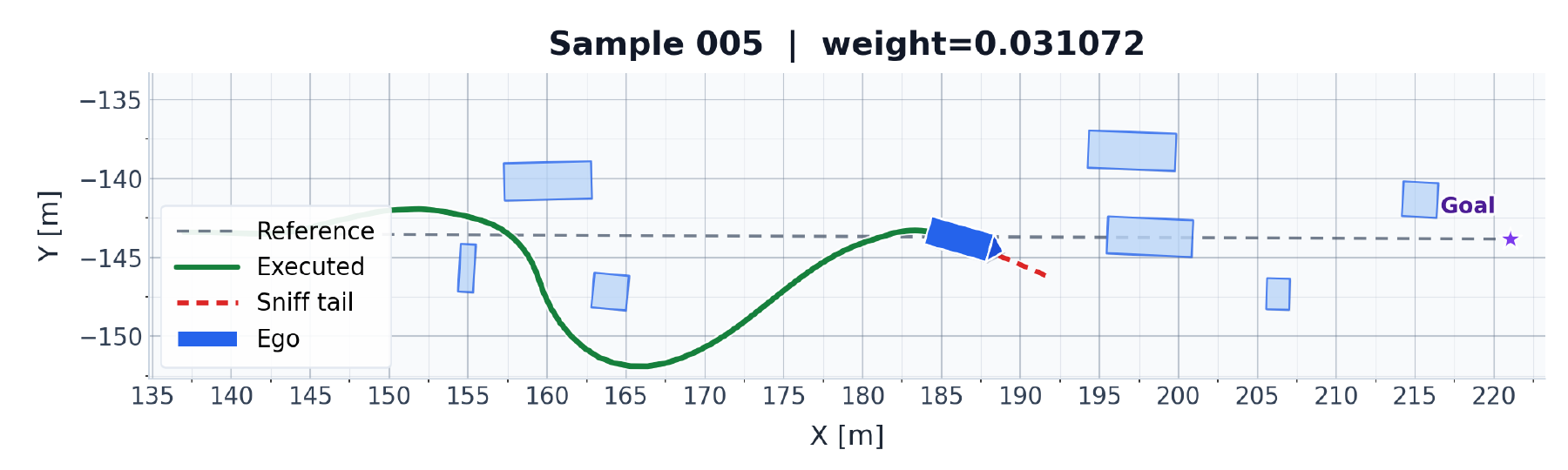} &
        \includegraphics[width=0.32\textwidth]{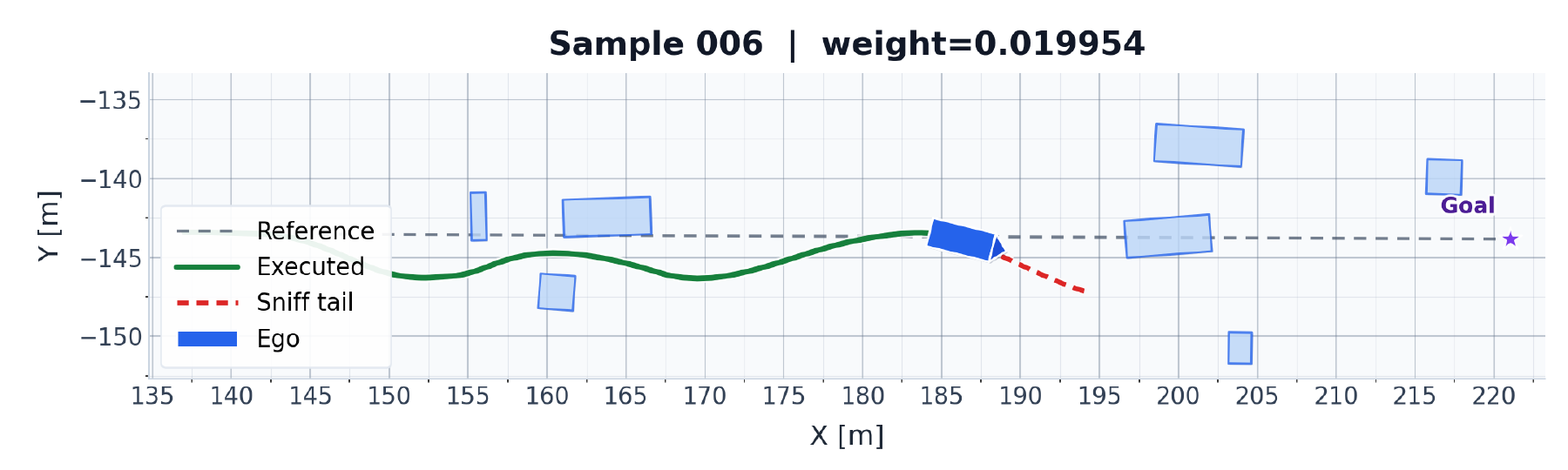}
    \end{tabular}
    \caption{Simulation-in-the-loop MPC rollouts under different plausible scene samples.}
    \label{fig:forward_sim}
    \vspace{-3mm}
\end{figure*}

To understand why SIGMA outperforms the baseline schemes, we analyze the planning-gain assessment process through forward simulation within the MPC framework. Instead of relying on perception uncertainty, we evaluate how different uncertainty realizations affect downstream planning outcomes. Specifically, we generate multiple scene samples using Monte Carlo sampling over obstacle uncertainty and perform MPC rollouts for each sampled scenario. Representative results are shown in Fig.~\ref{fig:forward_sim}, where each column corresponds to a different scene sample and the resulting MPC trajectory.

We observe that not all uncertainty realizations lead to meaningful changes in planning. In some cases, different scene samples result in nearly identical trajectories, indicating that resolving such uncertainty would not improve the final planning outcome. However, in critical regions (e.g., near obstacle clusters), different uncertainty realizations lead to significantly different feasible trajectories, directly affecting collision avoidance and path selection.

These results highlight that the impact of uncertainty should be evaluated in the planning space rather than the perception space. Only uncertainty that leads to divergent planning outcomes is decision-relevant. By explicitly evaluating this effect through forward simulation, SIGMA identifies uncertainty with significant expected planning gain and invokes cloud interaction only in decision-critical situations. This simulation-in-the-loop mechanism, as detailed in Section~\ref{subsec:mc_value}, explains why SIGMA achieves both improved navigation performance and reduced cloud usage compared to fixed-interval strategies.

\subsection{Experiment 3: More Quantitative Evaluations}

We evaluate the LOS, PCS, and SIGMA schemes at a lower speed (target speed of 3\,m/s) with a shorter horizon ($H=10$) and a smaller time step ($\Delta t=0.1$\,s) in more scenarios. By randomly removing some of the unknown objects in Fig.~\ref{figure:setting}, we configure different scenarios with $U\in\{0,1,2\}$. Quantitative results in each scenario are obtained by averaging $30$ random simulation runs, with independent realizations in each run. Note that the LOS and PCS schemes may fail due to collision or timeout. Here, we compute trajectory-quality metrics only over their successful cases, while failure cases are used to compute the success rates.

The evaluation results under different numbers of unknown objects are shown in Table~\ref{table:params}. As the number of unknown obstacles increases from $0$ to $2$, all schemes require more collision-avoidance maneuvers, leading to increased trajectory length or finish time. When there are no unknown objects, the three schemes show similar performance. With one unknown object, LOS becomes more conservative because it cannot recognize abnormal objects, while PCS benefits from fixed-period cloud assistance and achieves a slightly shorter finish time than SIGMA. However, when multiple unknown objects are present, fixed-period queries may occur at suboptimal times, which degrades PCS performance. In contrast, SIGMA uses cloud-side LVM reasoning to resolve decision-relevant unknown objects and relies on simulation-in-the-loop EPG estimation in Section~\ref{subsec:mc_value} to determine query timing. As a result, SIGMA achieves the shortest trajectory length in all tested settings and the shortest finish time in the most challenging case with $U=2$. Therefore, we do not claim that SIGMA is always the fastest; rather, it provides the largest efficiency gain when cloud-collaboration timing becomes decision-critical.

\begin{figure}[t]
    \centering
    \noindent
    \includegraphics[width=0.48\columnwidth]{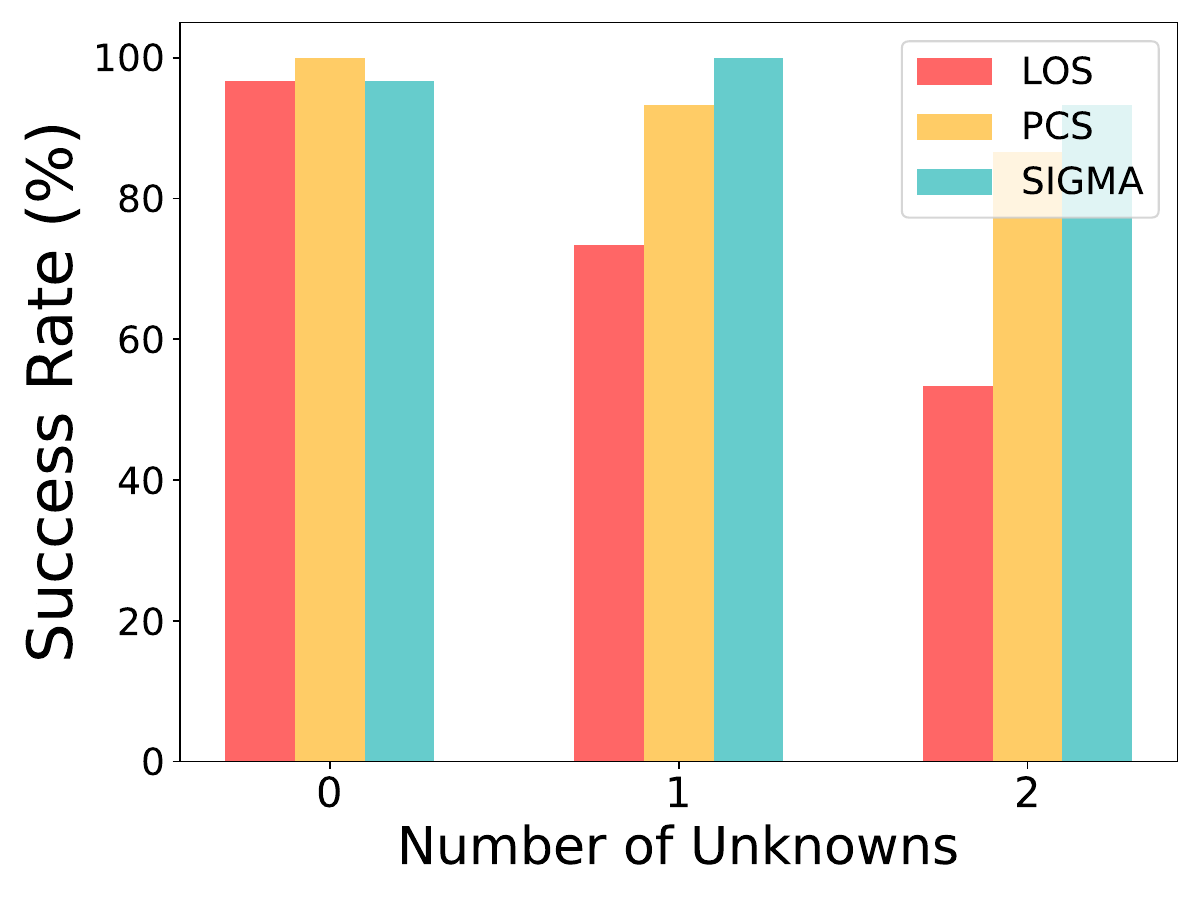}%
    \includegraphics[width=0.48\columnwidth]{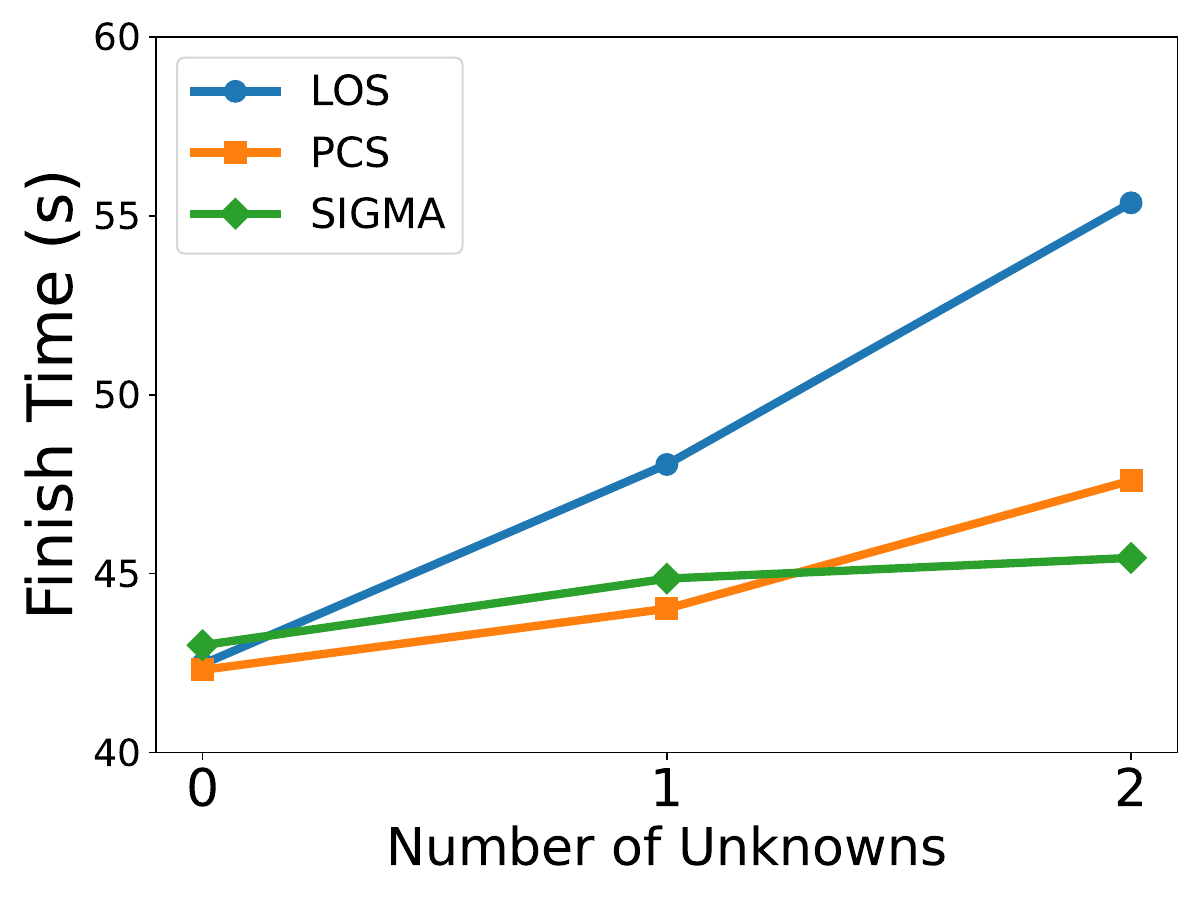}%
    \vspace{-2mm}
    \caption{Comparison of success rate and finish time in Experiment~3.}
    \label{figure:experiment1_success}
    \vspace{-2mm}
\end{figure}

Success rates are computed based on two failure conditions: 1) crash failure, where the ego vehicle collides with any obstacle or lane boundary; and 2) no-progress failure, where the ego vehicle gets stuck or moves off the route and fails to find a feasible path. As shown in Fig.~\ref{figure:experiment1_success}, SIGMA maintains a success rate close to $100\,\%$ in all scenarios. PCS works well under $0$--$1$ unknown objects, but causes safety issues under $2$ unknown objects, while LOS gives the lowest success rates. Compared with the second-best scheme, SIGMA improves the success rate by over $6\%$.
\begin{table}[!t]
\centering
\begin{center}
\vspace{-1mm}
\caption{Quantitative comparison of Experiment 3.}
\label{table:params}

\scalebox{1}{
\begin{tabular}{|c|c|c|c|c|c|c|}
\hline
\multirow{2}{*}{$U$} &
{\textbf{Method}} & {\textbf{FTime}} & {\textbf{TLen}} & {\textbf{AvgLD}} & {\textbf{SVar}} & {\textbf{MLat}} \\
                          &  (Unit)
                            & (s)                            & (m)                             & (m)                              & (\text{m/s})                   & (m) \\
\hline
0 & LOS  & 42.47 & 114.27 & 0.52 & 2.34 & 1.59 \\
0 & PCS  & 42.31 & 114.63 & 0.53 & 2.64 & 1.41 \\
0 & SIGMA & 43.00 & 113.56 & 0.51 & 2.41 & 1.62 \\
\hline
1 & LOS  & 48.05 & 119.95 & 0.49 & 2.57 & 1.54 \\
1 & PCS  & 44.02 & 114.44 & 0.53 & 2.68 & 1.43 \\
1 & SIGMA & 44.86 & 113.67 & 0.52 & 2.53 & 1.64 \\
\hline
2 & LOS  & 55.36 & 119.32 & 0.40 & 2.39 & 1.55 \\
2 & PCS  & 47.60 & 114.73 & 0.49 & 2.23 & 1.41 \\
2 & SIGMA & 45.44 & 112.79 & 0.54 & 2.48 & 1.51 \\
\hline
\end{tabular}
}
\end{center}
\vspace{-5mm}
\end{table}

\begin{figure}[!t]
    \centering
    \includegraphics[width=0.9\columnwidth]{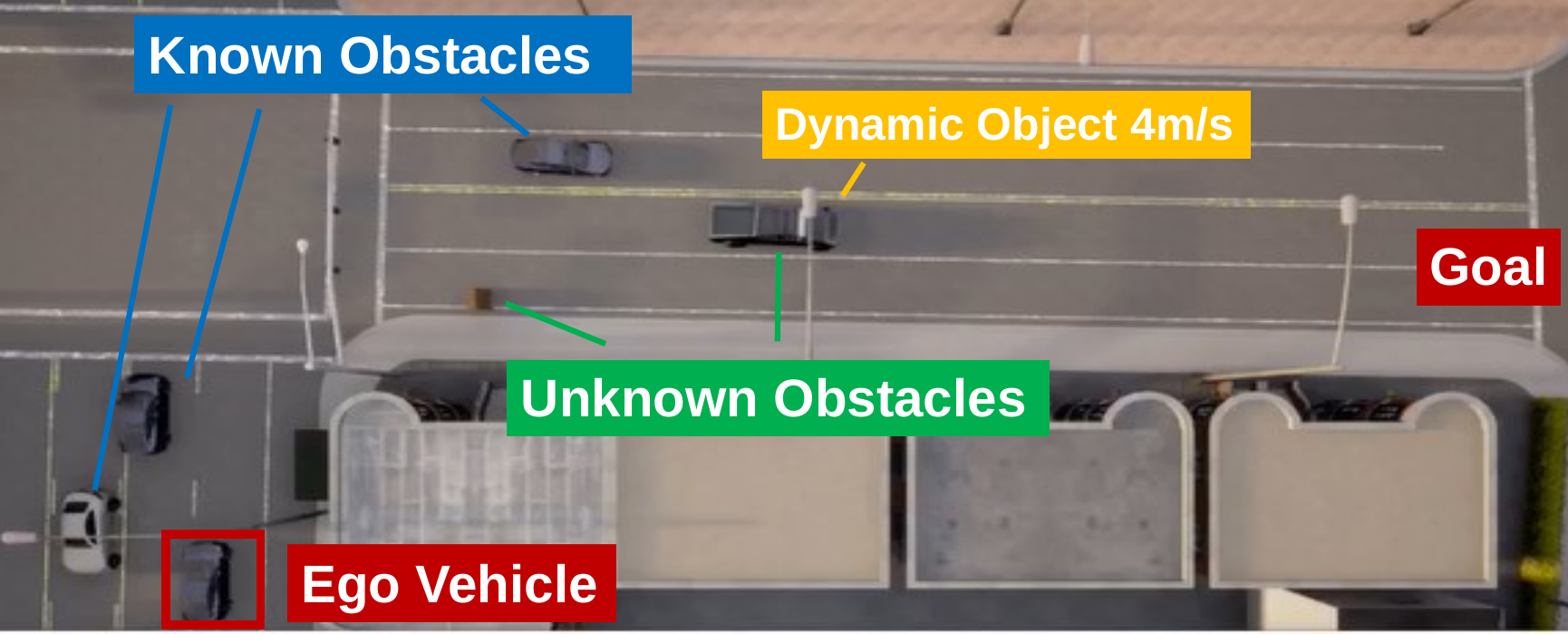}
    \vspace{-2mm}
    \caption{Dynamic-obstacle scenario configuration for Experiment~4.}
    \label{figure:setting_exp4}
    \vspace{-2mm}
\end{figure}

\subsection{Experiment 4: Scenarios with Dynamic Obstacles}

In this experiment, we evaluate the performance of the proposed framework in a dynamic-obstacle scenario at a crossroad in the CARLA Town03 map, as shown in Fig.~\ref{figure:setting_exp4}. The environment includes two unknown objects and three known objects. Among the unknown objects, one is a static doghouse and the other is a dynamic cybertruck moving at a constant speed of $4$ m/s. The LOS scheme always fails near the turning point due to uncertain perception of the doghouse and cybertruck, which overly restricts the feasible solution space of the MPC planner. This failure case illustrates a typical limitation of purely local uncertainty handling: when unknown objects appear in a narrow decision region, conservative obstacle treatment may make the planner unable to find a feasible and efficient maneuver. Therefore, we only compare PCS and SIGMA in this experiment.

\begin{figure}[!t]
    \vspace{-1mm}
    \centering
    \subfloat[PCS.]{
        \includegraphics[width=0.45\columnwidth]{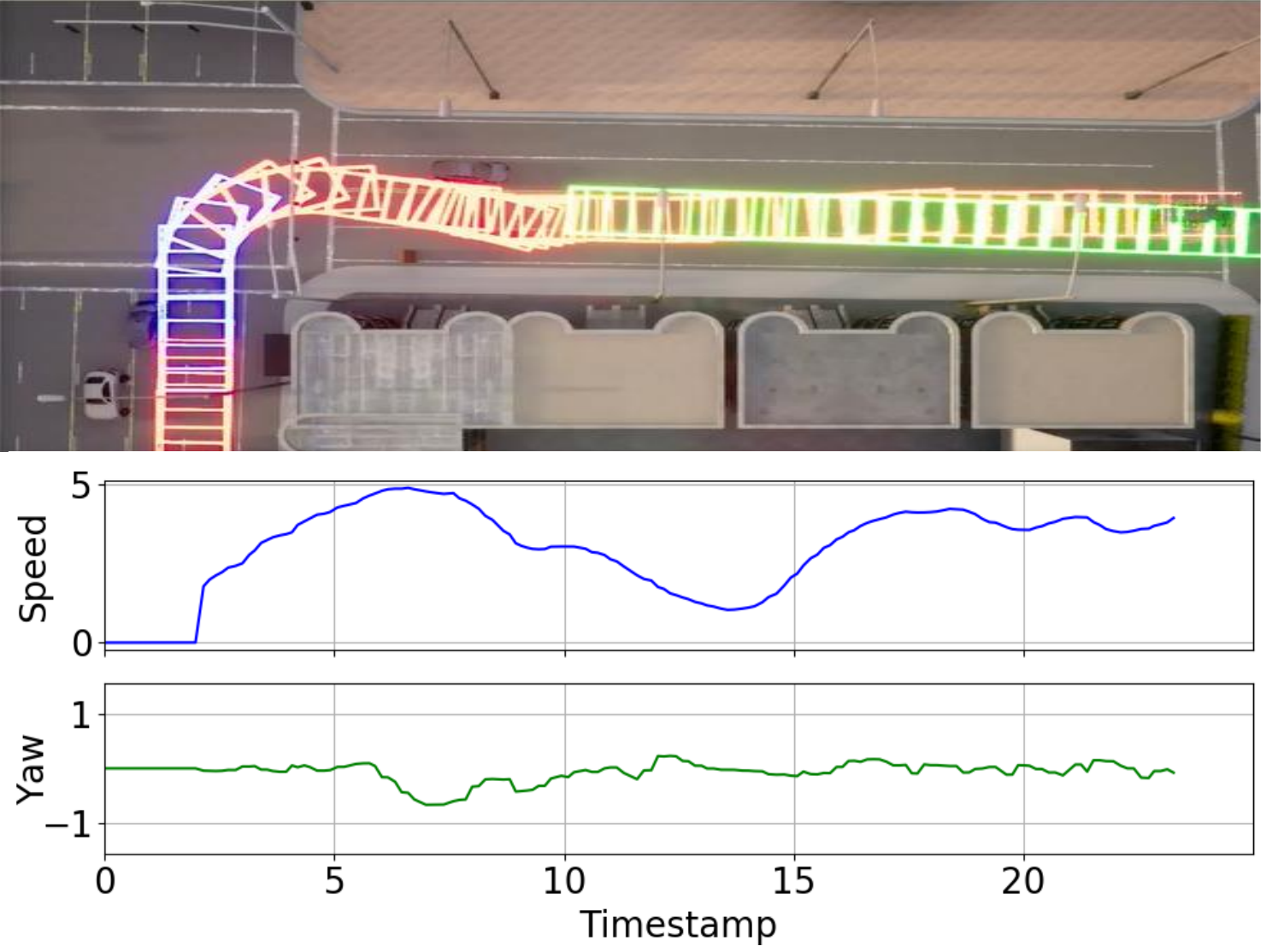}
        \label{fig:exp4_pcs}
    }
    \hfil
    \subfloat[SIGMA.]{
        \includegraphics[width=0.45\columnwidth]{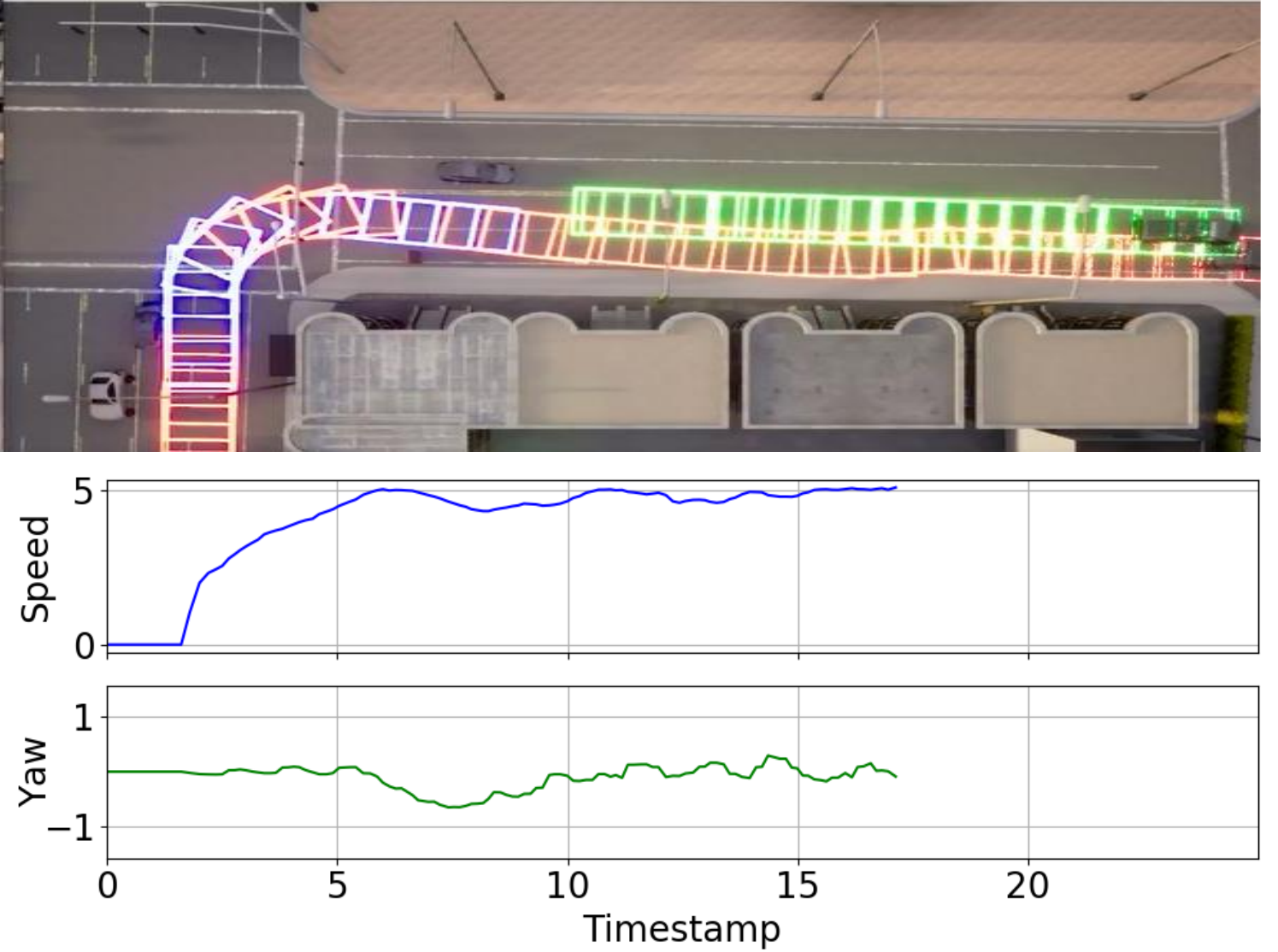}
        \label{fig:exp4_sigma}
    }
    \caption{Trajectory and control profiles in Experiment~4.}
    \label{fig:experiment4}
    \vspace{-2mm}
\end{figure}

For PCS, the ego vehicle turns right while avoiding collision with the doghouse, and then follows the lead vehicle cybertruck until reaching the goal. This behavior is safe but conservative, since the fixed-period collaboration strategy does not necessarily query the cloud when the cybertruck becomes decision-critical. In contrast, under SIGMA, the ego vehicle triggers another cloud query during the car-following phase behind the cybertruck. This is because the simulation-in-the-loop EPG estimation mechanism (Section~\ref{subsec:mc_value}) identifies that resolving the current uncertainty can still change the downstream planning outcome. In other words, the remaining uncertainty is not only a perception ambiguity, but also a maneuver-level ambiguity between continuing to follow the lead vehicle and overtaking it. The accepted cloud query therefore preserves the opportunity for subsequent collaboration and enables the ego vehicle to overtake the cybertruck by following cloud-generated overtaking waypoints rather than local car-following waypoints, as shown in Fig.~\ref{fig:experiment4} (red boxes denote the local controller, blue boxes denote LVM-guided MPC, and green boxes denote the dynamic obstacle).

Consequently, compared with PCS, the proposed SIGMA completes the trip in a significantly shorter time, achieving more than 26.2\% reduction in finish time. It is also worth noting that SIGMA does not simply minimize trajectory length; instead, it accepts a slightly longer maneuver when it leads to a better time-efficient decision. This reflects the key advantage of planning-gain-guided collaboration in dynamic scenarios: cloud reasoning is invoked when it can change the driving strategy, rather than merely refine perception. Detailed quantitative results are shown in Table~\ref{table:exp4}.

\begin{table}[t]
\centering
\caption{Quantitative comparison in Experiment~4.}
\label{table:exp4}
\setlength{\tabcolsep}{4pt}
\renewcommand{\arraystretch}{1.05}
\scalebox{1.0}{
\begin{tabular}{lccccc}
\toprule
\textbf{Method} & \textbf{FTime(s)} & \textbf{TLen(m)} & \textbf{AvgLD(m)} & \textbf{SVar(m/s)} & \textbf{MLat(m)} \\
\midrule
PCS   & 23.40 & 71.47 & 0.47 & 2.99 & 1.68 \\
SIGMA & 17.28 & 72.90 & 0.62 & 2.29 & 1.44 \\
\bottomrule
\end{tabular}
}
\vspace{-2mm}
\end{table}

\section{Conclusion}

This paper presented \textbf{SIGMA}, a simulation-in-the-loop planning-gain assessment framework for fast--slow collaborative autonomous driving. Instead of relying on perception-level uncertainty or heuristic triggers, SIGMA evaluates uncertainty through its impact on planning outcomes via simulation-in-the-loop forward simulation. We introduced EPG to quantify the expected improvement in planning performance if uncertainty is resolved, which serves as a unified criterion for cloud invocation, cloud-guidance integration, and resource allocation. Semantic uncertainty is modeled as a confidence-based screening signal for unknown-object awareness, while geometric uncertainty is represented probabilistically and propagated into motion planning through chance-constrained obstacle inflation. Under limited cloud resources, an EPG- and deadline-aware allocation mechanism ensures that cloud assistance is provided only when it is both decision-relevant and timely. Experiments in CARLA demonstrate that the proposed framework reduces unnecessary cloud usage while improving planning performance, driving efficiency, and success rates in both static and dynamic obstacle scenarios.

Future work will extend SIGMA along three directions. First, we will move beyond static or weakly dynamic obstacle uncertainty and model interaction-aware uncertainty in dynamic driving scenarios, including multi-modal agent intentions and trajectory prediction ambiguity. Second, we will replace the current threshold-based cloud invocation and resource allocation rules with reinforcement-learning-based policies that optimize long-term planning performance under latency, cost, and safety constraints. Third, we will upgrade the current simulation-in-the-loop EPG estimator into a world-model-in-the-loop framework, where generative world models provide realistic counterfactual rollouts for evaluating the value of cloud reasoning in complex interactive environments. \rev{More broadly, this direction connects SIGMA to the physical-AI agent-harness perspective, where simulation, verification, safety checking, and execution monitoring wrap model reasoning before actions are deployed in the physical world.}

\bibliographystyle{IEEEtran}
\bibliography{main}
\end{document}